\documentclass{article} 
\usepackage{colm2024_conference}

\usepackage{booktabs}
\usepackage{graphicx}
\usepackage{enumitem}
\usepackage{changes}
\usepackage{wrapfig}
\usepackage{algorithm}
\usepackage{algpseudocode}
\usepackage{natbib}
\usepackage{makecell}
\usepackage{booktabs}
\usepackage{array}
\usepackage{amsmath}
\usepackage{amssymb}
\usepackage{amsfonts}
\usepackage{threeparttable}
\usepackage{multirow}
\usepackage{verbatim}
\usepackage{caption}
\usepackage{longtable}
\usepackage{makecell}
\usepackage{ragged2e}      
\usepackage{array}         
\usepackage{supertabular}
\usepackage{CJKutf8}
\usepackage{array} 
\usepackage[utf8]{inputenc} 
\usepackage[T1]{fontenc}
\usepackage[french,vietnamese,mongolian,greek,english]{babel}
\usepackage{pifont}
\usepackage{xcolor}
\usepackage{enumitem}
\usepackage{tablefootnote}
\usepackage{xspace}
\usepackage{textcomp}
\usepackage{makecell}
\usepackage{lscape} 
\usepackage{siunitx}
\usepackage{listings}
\usepackage{xcolor}
\usepackage{colortbl}
\usepackage{ulem}
\usepackage{amsmath}
\usepackage{multirow, array, graphicx} 

\hypersetup{
    colorlinks=true
}
\definecolor{dt}{gray}{0.7}
\usepackage{pifont}       
\usepackage{bbding}      
\usepackage{fontawesome}
\newcolumntype{L}[1]{>{\raggedright\arraybackslash}m{#1}}

\usepackage{scrextend}

\usepackage{tgpagella}
\usepackage{latexsym}
\usepackage[T1]{fontenc}
\usepackage{microtype}
\definecolor{mydarkblue}{rgb}{0,0.08,0.45}
\definecolor{citecolor}{HTML}{0071BC}
\usepackage{url}            
\usepackage{nicefrac}       
\usepackage{changepage}
\usepackage{xargs}          
\usepackage{wrapfig,lipsum,booktabs}
\usepackage{longtable}
\usepackage{subcaption}
\usepackage{endnotes}

\usepackage{fancyvrb}
\usepackage{fvextra}
\usepackage{pgfplots}
\usetikzlibrary{pgfplots.groupplots}
\pgfplotsset{compat=1.3}
\usepackage{tikz}
\usetikzlibrary{patterns}

\usepackage[most]{tcolorbox}
\usepackage[capitalize,noabbrev]{cleveref}
\crefname{section}{Section}{\S\S}
\Crefname{section}{Section}{\S\S}
\crefname{table}{Table}{Tables}
\crefname{figure}{Figure}{Figures}
\crefname{algorithm}{Algorithm}{}
\crefname{equation}{eq.}{}
\crefname{appendix}{Appendix}{}
\crefformat{section}{Section #2#1#3}
\usepackage{multicol}
\usepackage{tcolorbox}

\usepackage{titlesec}
\titleformat*{\section}{\large\bfseries}

\title{SAIL-VL2 Technical Report}

\author{
\bf Douyin SAIL Team, \bf LV-NUS Lab}

\begin{document}

\maketitle

\begin{abstract}
We introduce SAIL-VL2, an open-suite vision-language foundation model (LVM) for comprehensive multimodal understanding and reasoning. As the successor to SAIL-VL, SAIL-VL2 achieves state-of-the-art performance at the 2B and 8B parameter scales across diverse image and video benchmarks, demonstrating strong capabilities from fine-grained perception to complex reasoning. Its effectiveness is driven by three core innovations. First, a large-scale data curation pipeline with scoring and filtering strategies enhances both quality and distribution across captioning, OCR, QA, and video data, improving training efficiency. Second, a progressive training framework begins with a powerful pre-trained vision encoder (SAIL-ViT), advances through multimodal pre-training, and culminates in a thinking-fusion SFT–RL hybrid paradigm that systematically strengthens model capabilities. Third, architectural advances extend beyond dense LLMs to efficient sparse Mixture-of-Experts (MoE) designs. With these contributions, SAIL-VL2 demonstrates competitive performance across 106 datasets and achieves state-of-the-art results on challenging reasoning benchmarks such as MMMU and MathVista. Furthermore, on the OpenCompass leaderboard, SAIL-VL2-2B ranks first among officially released open-source models under the 4B parameter scale, while serving as an efficient and extensible foundation for the open-source multimodal community.
\end{abstract}

\begin{figure}[h]
    \centering
    
    \vspace{-3mm}\includegraphics[width=1\linewidth]{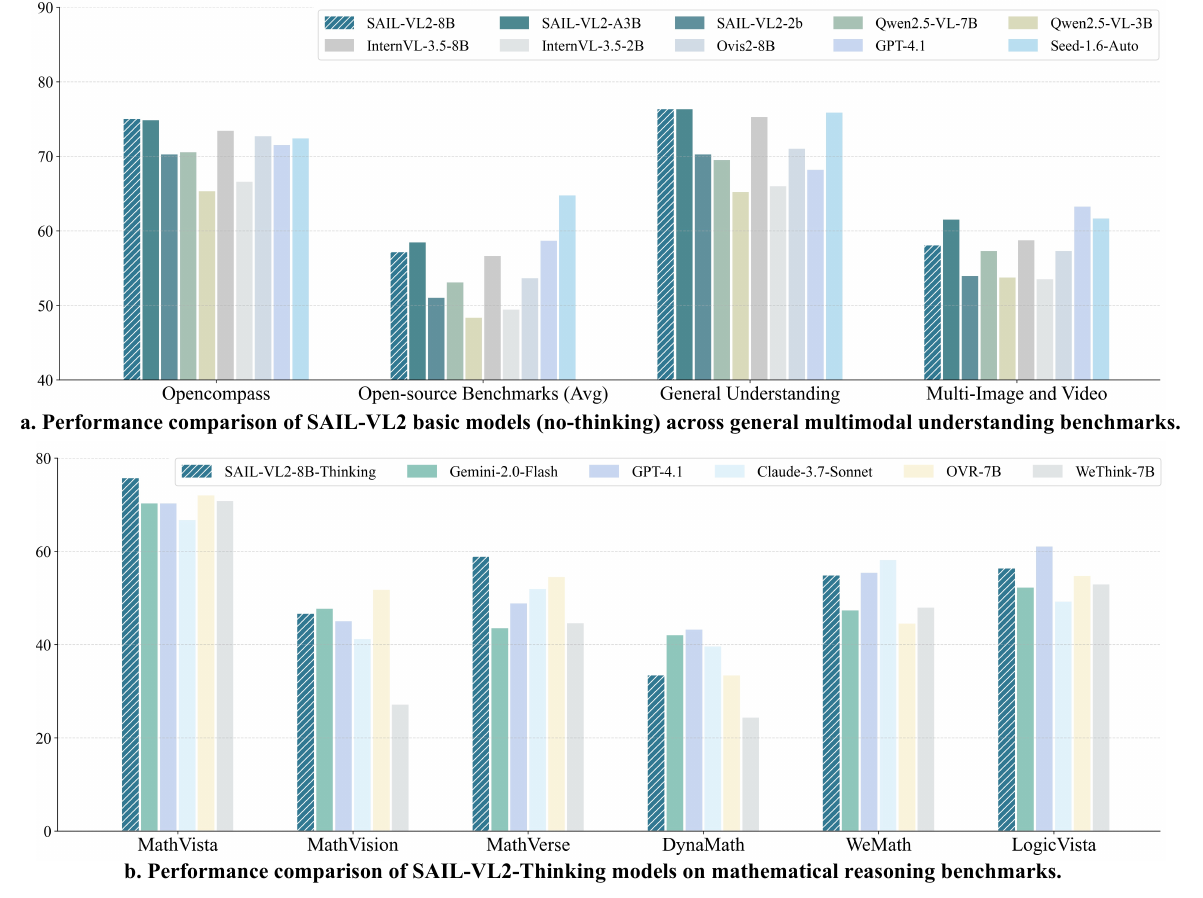}
    \vspace{-5mm}
    \caption{Performance comparison between \textbf{SAIL-VL2 (2B/8B/A3B)} and other LVMs (both open-source and large-scale closed-source). SAIL-VL2 demonstrates strong performance across multiple dimensions.}
    \label{fig:fig1}
\end{figure}
\section{Introduction}
Large-scale Vision-Language Models (LVMs)~\citep{qwen2_5_vl,internvl,internvl3,mimovl,kimi_vl} bridge the gap between vision and language by integrating visual representations with linguistic descriptions in a shared semantic space. This shift from unimodal to cross-modal and multimodal understanding mirrors natural human interactions with the world. Driven by advances in large language models (LLMs)~\citep{brown2020language,llama,qwen3} and visual representation techniques~\citep{radford2021learning,li2023blip}, LVMs have evolved from coarse-grained visual understanding to fine-grained multimodal reasoning. Concurrently, training paradigms have progressed from teacher-forcing supervised learning to hybrid methods, integrating supervised fine-tuning (SFT) and reinforcement learning (RL)~\citep{ouyang2022training,grpo,yu2025dapo} for self-improvement. 
With continuous advancements in technology and scaling-up training data and model parameters, LVMs are steadily advancing toward Artificial General Intelligence (AGI).

Scaling up model parameters and training data to make LVMs 'larger' has emerged as a pivotal approach for pushing the performance boundaries of LVMs~\citep{seed_1_5_vl,qwen2_5_vl,internvl3}.  While this paradigm has yielded substantial performance gains, it also imposes considerable challenges in terms of computational demands and training, as well as deployment costs.
In contrast, our SAIL-VL series focuses on developing efficient LVMs, aiming to explore \textit{'how knowledge can be effectively injected through efficient architectures and training strategies,'} thereby establishing an open-source model family that embodies the principle of \textit{'small model, strong performance'}. 

To advance the development of more powerful yet efficient LVMs, we present \textbf{SAIL-VL2}, the latest iteration of our research, building upon our previous work, SAIL-VL~\citep{dong2025scalable}. SAIL-VL2 introduces substantial upgrades in architecture, training strategies, and data quality, achieving state-of-the-art performance across diverse benchmarks at a comparable parameter scale. These advancements endow the model with superior capabilities, ranging from multimodal understanding to complex reasoning.

From the data perspective, we design a comprehensive scoring and filtering pipeline that covers the full spectrum of multimodal inputs, ranging from captioning to QA and from images to videos. This pipeline systematically enhances both quality and distribution, thereby improving data efficiency across pre-training and post-training stages. In terms of training, we develop a progressive and efficient framework: beginning with a powerful pre-trained vision encoder (SAIL-ViT), advancing through basic multimodal pre-training, and culminating in a thinking-fusion SFT–RL hybrid paradigm, which enables systematic capability enhancement. Architecturally, SAIL-VL2 moves beyond conventional dense LLMs by adopting more efficient sparse Mixture-of-Experts (MoE) designs.

With the comprehensive upgrades and designs described above, the core capabilities and highlights of SAIL-VL2 can be summarized as follows~(Figure~\ref{fig:fig1}):
\begin{itemize}
    \item \textbf{SAIL-VL2 is powerful yet efficient:}
    With training on 776B tokens, SAIL-VL2 has verified its effectiveness across 106 datasets, achieving state-of-the-art results on a broad spectrum of influential benchmarks under the 2B-parameter scale.
    Remarkably, even without specialized prompting, the base SAIL-VL2 model delivers highly competitive performance on challenging reasoning benchmarks such as MMMU and MathVista, demonstrating strong out-of-the-box capabilities.

    \item \textbf{SAIL-VL2 as a deep thinker:} Many real-world tasks demand sophisticated reasoning and multi-step thought processes, which remain challenging for standard LVMs. To address this, we develop \textit{SAIL-VL2-Thinking}, a specialized variant trained with advanced Chain-of-Thought (CoT) and reinforcement learning (RL) strategies. This design substantially improves performance on complex reasoning benchmarks, often matching or even surpassing models with far larger parameter scales, thereby setting a new standard for efficient architectures in high-level reasoning.

   \item \textbf{SAIL-VL2 perceives with clarity:} Fine-grained visual understanding is a critical challenge for multimodal models. SAIL-VL2 delivers high-fidelity perception in tasks such as OCR, high-resolution document layout analysis, and complex chart interpretation, achieving detailed visual grounding beyond models of similar scale.
\end{itemize}

In summary, SAIL-VL2 represents a comprehensive advancement in the design of efficient large vision-language models, integrating innovations in architecture, training strategies, and data curation. To foster openness and collaboration, we will release the full SAIL-VL2 model suite along with its inference code. We envision SAIL-VL2 as an efficient and extensible foundation that not only advances state-of-the-art performance but also empowers the broader open-source multimodal ecosystem.
\section{Model Architecture} 
\begin{figure}[t]
    \centering
    \includegraphics[width=1\linewidth]{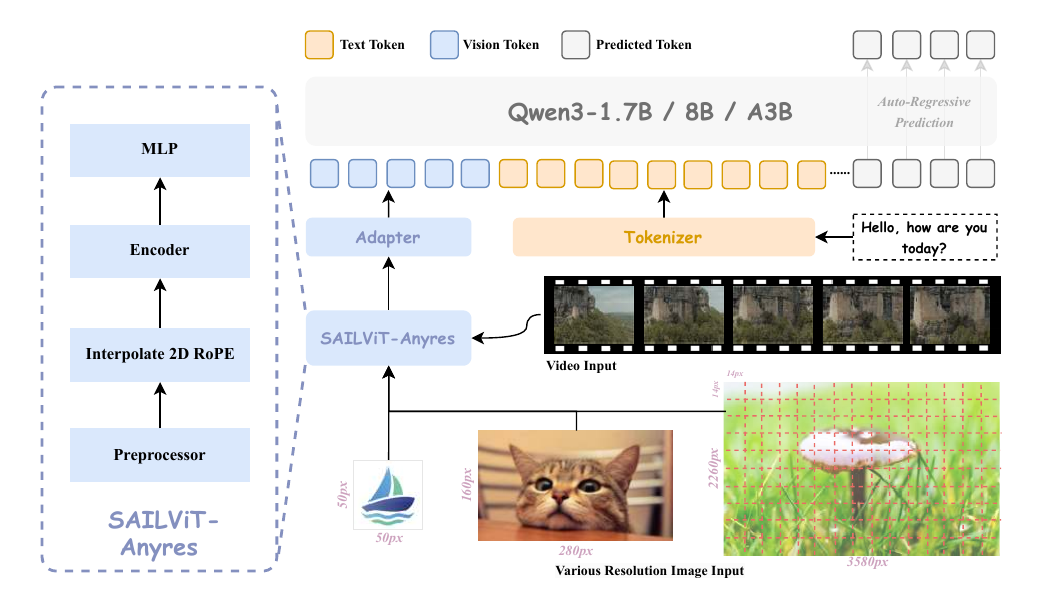}
    \caption{\textbf{Overview of the SAIL-VL2 framework.} The architecture is composed of a vision encoder (SAIL-ViT) that aligns visual inputs into the representation space of the LLM. A lightweight adapter further transforms visual embeddings into tokenized representations, which are jointly processed with linguistic embeddings for multimodal reasoning and prediction. SAIL-VL2 accommodates multiple LLM backbones, ensuring flexibility and scalability across model configurations.
    }
    \label{fig:framework}
\end{figure}

As shown in Table~\ref{tab:sail-vl2-arch} and Figure~\ref{fig:framework}, we initialize the LLM with the Qwen3 series models and the ViT with the SAIL-ViT series models. In this section, we introduce the architecture of SAIL-VL2, which follows the general framework of LVMs and is composed of three core components:
\begin{table}[h]
\centering
\resizebox{\textwidth}{!}{%
\renewcommand{\arraystretch}{1.3}
\begin{tabular}{lllcclll}
    \toprule
    \textbf{Model} & \textbf{Vision Encoder} & \textbf{Language Model} & \multicolumn{3}{c}{\textbf{\#Param}} \\
    \cmidrule(lr){4-6}
    & & & Vision & Language & Total \\
    \midrule
    \rowcolor{gray!15} \multicolumn{6}{c}{\textit{Dense Models}} \\
    SAIL-VL2-2B        & SAIL-ViT-Huge        & Qwen3-1.7B   & 0.6B & 2.0B   & 2.6B  \\
    SAIL-VL2-8B        & SAIL-ViT-Huge        & Qwen3-8B     & 0.6B & 8.2B   & 8.8B  \\
    \midrule
    \rowcolor{gray!15} \multicolumn{6}{c}{\textit{MoE Models}} \\
    SAIL-VL2-30B-A3B      & SAIL-ViT-Huge & Qwen3-30B-A3B      & 0.6B & 30.5B   & 31.1B (A3B)  \\
    \bottomrule
\end{tabular}
}
\caption{
Configurations of different scales and variants in our SAIL-VL2 series.}
\label{tab:sail-vl2-arch}
\end{table}

\textbf{Vision Encoder:} The vision encoder of SAIL-VL2 builds on the Vision Transformer (ViT) architecture, which encodes images and videos into sequences of visual tokens for multimodal reasoning within the LLM. The core challenge lies in the modality and semantic gap, as visual and linguistic tokens originate from heterogeneous spaces and must be aligned into a unified representation. To address this, we introduce \textit{SAIL-ViT}~\citep{yin2025sailvit}, a tailored vision encoder that evolves from ViT through a progressive training pipeline, enabling step-by-step alignment of visual features with the LLM’s representation space. The system encompasses a family of visual encoders, including a standard image encoder, and an any-resolution encoder, thereby accommodating diverse multimodal requirements. This progressive design enhances visual–linguistic alignment in a stepwise manner and ultimately achieves comprehensive cross-modal integration.

\textbf{Vision-Language Adapter:} 
The vision-language adapter is a lightweight two-layer MLP that projects the output of SAIL-ViT into the language space, adjusting dimensions and mitigating the modality gap.

\textbf{Large Language Models:} 
For the foundation language models in SAIL-VL2, we explore both dense LLMs (Qwen3-Instruct~\citep{qwen3} series) and sparse MoE architectures (Qwen3-MoE~\citep{qwen3}). Given an instruction, it is first embedded into a sequence of linguistic tokens, which are then processed jointly with the aligned visual tokens in the LLM, enabling unified multimodal understanding.

\subsection{SAIL-ViT}
The vision encoder is a core component for perceiving visual information and aligning it with the linguistic space. To this end, we develop SAIL-ViT, a pre-trained vision encoder. SAIl-ViT is progressively optimized through a training strategy that injects multi-granularity knowledge, enabling comprehensive alignment with LLMs.

\subsubsection{Progressive Training Strategy}

We first introduce a three-stage progressive training strategy that incrementally aligns the vision encoder with the LLM’s representation space by injecting knowledge of different granularities and leveraging corresponding training data. All three stages are trained in an instruction-tuning manner. We next describe each stage in detail:

\textbf{Stage I: Warm-up adaptation.} In this stage, both the vision encoder and the LLM are frozen, and only the parameters of the adapter are tuned. The goal is to activate the adapter’s capacity to perform a coarse-grained adaptation of the vision encoder outputs into the LLM domain. For training, we employ 8M simple multimodal understanding examples, including 4.9M captioning samples from the SAIL-Caption dataset~\citep{dong2025scalable} and 3.1M OCR samples from the IDL-WDS dataset~\citep{biten2022ocr}. Training is conducted for one epoch with a learning rate of $2 \times 10^{-4}$  and a batch size of 1920.

\textbf{Stage II: Fine-grained alignment.} Here, we keep the LLM fixed while unlocking both the vision encoder and the adapter, enabling deeper and more comprehensive alignment. Beyond the datasets used in Stage I, we expand the scale and diversity of captioning data with an additional 6.7M samples from SAIL-Caption, supplement the OCR source with DocStruct~\citep{wang2020docstruct}, and further incorporate video-caption data for joint training, thereby enriching distributional diversity. This stage is trained with a reduced learning rate of $2 \times 10^{-5}$  and a batch size of 512.

\textbf{Stage III: World knowledge injection.} To move beyond conventional alignment, we introduce a stage that integrates broad multimodal understanding and reasoning tasks, enhancing the vision encoder’s generalization and enabling comprehensive alignment with the LLM across multiple levels. In this stage, we unlock all parameters, including the vision encoder, adapter, and LLM, and train on a wide spectrum of data (36.5M): captioning~\citep{deitke2024molmo,dong2025scalable}, OCR~\citep{mathew2021docvqadatasetvqadocument,laurençon2024buildingbetterunderstandingvisionlanguage,wang2020docstruct,deitke2024molmo}, open-ended QA~\citep{chen2023sharegpt4vimprovinglargemultimodal,cui2025comprehensive}, math reasoning~\citep{shi2024mathllavabootstrappingmathematicalreasoning,wang2025vgr,cao2022augmented,amini2019mathqainterpretablemathword}, short QA~\citep{biten2019scenetextvisualquestion,goyal2017makingvvqamatter,marino2019okvqavisualquestionanswering}, and pure text corpora~\citep{OpenHermes2.5,xu2024magpiealignmentdatasynthesis}. This diverse and large-scale training significantly boosts the generalization ability of the vision encoder across tasks. Training is performed for one epoch with a learning rate of 1e-5 and a batch size of 512.

The vision encoder trained in the final stage is taken as SAIL-ViT and provides the foundation for subsequent SAIL-VL2 training.

\subsubsection{SAIL-ViT Family}

We employ AIMv2-Large/Huge~\citep{fini2024multimodal} as the foundation vision encoder and instantiate SAIL-ViT with two variants, further optimized under our progressive training schedule. The first, \textit{vanilla ViT}, accepts fixed 448×448 inputs. Each image is split into 32×32 non-overlapping patches (patch size 14), producing 1,024 tokens fused with learnable positional embeddings. High-resolution images are tiled into multiple 448×448 crops, encoded independently, and their tokens aggregated for downstream processing. The second, \textit{SAIL-ViT-AnyRes}, supports arbitrary resolutions. Since conventional positional embeddings are fixed-length and hard to generalize, we adopt an interpolation-based mechanism: pre-trained embeddings are resized to match the input resolution, providing a global prior that improves extrapolation. Images are thus converted into token sequences proportional to their resolution.

We conducted large-scale pre-training on both proposed SAIL-ViT variants, building powerful vision encoders that deliver strong visual representations.

\subsection{Multi-modal Mixture-of-Experts}
Mixture-of-Experts (MoE)~\citep{kimi_vl,li2024aria,shu2024llava,li2025uni,hong2025glm,huang2024mixture} provides an effective strategy for scaling LLMs by replacing standard MLP layers with parallel expert modules. A gating function activates only a small subset of experts for each token, enabling parameter scaling while preserving computational efficiency through sparse activation. In SAIL-VL2, we adopt Qwen3-based MoE structures and introduce strategies to ensure both stability and scalability. Balanced expert activation is critical for consistent scaling. Following prior work, we employ an auxiliary load-balancing loss and average activation across ranks to improve stability. Beyond loss design, we observe that activation patterns vary with data distribution, underscoring the importance of distribution-aware tuning. To preserve expert specialization learned during pre-training, we conduct data probing and automatic calibration on language data, which maximizes expert activation entropy. This maintains activation behavior on text-only calibration sets and significantly improves entropy on multimodal sets.

SAIL-VL2 is ultimately composed of a well-trained SAIL-ViT, an adapter, and an LLM. In the following, we present the comprehensive training strategy that empowers this architecture to achieve effective multimodal alignment and strong vision-language reasoning capabilities.
\section{Pre-Training}

In this section, we detail the pre-training process of SAIL-VL2, including the curation and organization of pre-training data as well as the training recipe.

\subsection{Pre-Traning Data}
Building on the pre-training data of SAIL-VL~\citep{dong2025scalable}, SAIL-VL2 further upgrades and expands SAIL-Caption into SAIL-Caption2, providing more accurate annotations and a broader data distribution.
Beyond basic caption data, we also construct a generic QA corpus through a Caption2QA procedure. Specifically, a subset of captions from SAIL-Caption2 is transformed into QA pairs using a powerful LVM guided by carefully designed prompts. This process enriches the distributional diversity of QA data and enhances the model’s generalization capability.
In addition, following the data composition strategy of SAIL-VL, we incorporate pure text data~\citep{gu2024infinity,OpenHermes2.5,toshniwal2024openmath2} to preserve the LLM’s fundamental language modeling and comprehension ability during multimodal pre-training. We also collect large-scale multimodal instruction-tuning VQA datasets~\citep{xu2024llava,wang2024enhancing,laurenccon2024matters,tong2024cambrian1fullyopenvisioncentric,gu2024infinity,li2024llava}, which further enhance the model’s multimodal understanding and instruction-following capacity.

\begin{figure}[t]
    \centering
    \includegraphics[width=1.0\linewidth]{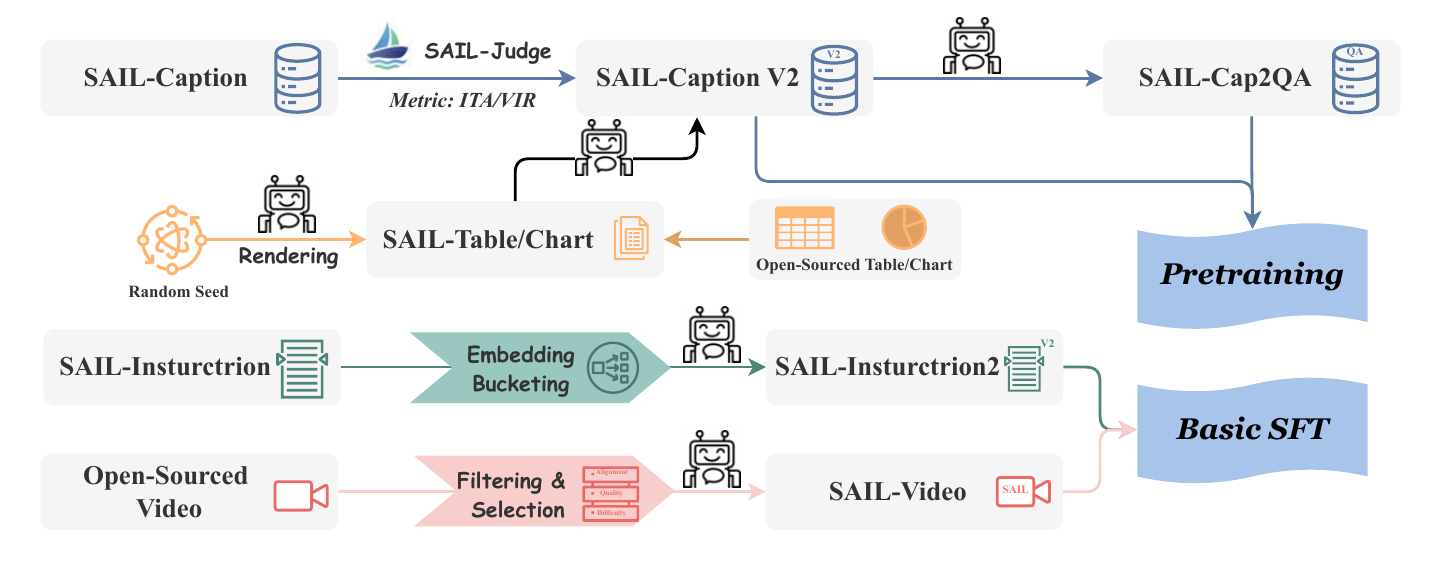}
    \caption{\textbf{Data construction pipeline for SAIL-VL2 training.} High-quality multimodal corpora are constructed by curating and filtering open-source datasets, and generating synthetic data, with both components systematically organized to meet the requirements of different training stages.
    }
    \label{fig:datapipeline}
\end{figure}

\subsubsection{SAIL-Caption2}
We upgrade SAIL-Caption2 in two aspects.
First, we introduce a caption judge model that provides comprehensive quality scoring, based on which low-quality samples are filtered out to improve dataset reliability.
Second, we broaden the diversity of captions by incorporating chart-related data, including both newly annotated chart captions and code-generated descriptions.

\textbf{Automatic Caption Quality Assessment and Filtering.} We assess caption quality from two perspectives: Visual Information Richness (VIR) and Image-Text Alignment (ITA).
\begin{itemize}
    \item VIR evaluates the informativeness of visual content across three aspects: (a) Instances \& Entities: number, diversity, and distinctiveness; (b) Visual Complexity: color variety, composition, and texture; (c) Fine-Grained Details: attributes such as textures, small objects, or contextual relationships.
    \item ITA measures how well captions match visual content across: (a) Specificity: precision of details; (b) Completeness: coverage of all key elements; (c) Accuracy: absence of hallucinations or redundancy.
\end{itemize}
For each evaluation dimension, captions are rated on a 1--5 scale, where higher scores indicate superior quality. We sample approximately 3M captions from \textit{SAIL-Caption} and obtain scoring annotations using the powerful LVM API. Our analysis reveals that 15--20\% of captions fall below an acceptable threshold (score $<3$), highlighting substantial noise in the raw corpus.  Directly relying on the LVM API to clean the entire dataset would incur prohibitive costs. Therefore, we train two judge models on \textit{SAIL-VL-2B}: a \textbf{Score Judge} (1--5 rating) and a \textbf{Yes-or-No Judge} (binary quality decision). To support robust judge model training, we construct a balanced dataset by uniformly resampling $\sim$500K captions across the full score distribution. Both judge models achieve over $90\%$ accuracy and recall (Table~\ref{tab:judge-results}). Notably, the base model SAIL-VL-1.5 already outperforms comparable zero-shot baselines such as InternVL3, demonstrating the strength and reliability of our judge model foundation. We then clean the dataset by removing captions with scores $<3$ or judged as ``No.'' Given judge accuracy above 90\%, the retained captions are estimated to reach $>$99\% high quality.
\begin{table}[h]
\centering
\renewcommand\arraystretch{1.3}
\begin{tabular}{lcc|lcc}
\hline
\multicolumn{3}{c}{\textbf{ITA Judge Model}} & \multicolumn{3}{c}{\textbf{VIR Judge Model}} \\
\hline
\textbf{Model} & \textbf{Precision} & \textbf{Recall} & \textbf{Model} & \textbf{Precision} & \textbf{Recall} \\
\hline
\rowcolor{gray!15} \multicolumn{6}{c}{\textit{Fine-tuned}} \\
\hline
SAIL-VL-1.5-2B-ITA-Score & 0.916 & 0.903 & \textbf{SAIL-VL-1.5-2B-VIR-Score} & 0.983 & 0.975 \\
\textbf{SAIL-VL-1.5-2B-ITA-YorN}  & 0.906 & 0.924 & SAIL-VL-1.5-2B-VIR-YorN  & 0.934 & 0.954 \\
SAIL-VL-1.5-8B-ITA-Score & 0.927 & 0.899 & SAIL-VL-1.5-8B-VIR-Score & 0.982 & 0.976 \\
SAIL-VL-1.5-8B-ITA-YorN  & 0.916 & 0.919 & SAIL-VL-1.5-8B-VIR-YorN  & 0.928 & 0.961 \\
\hline
\rowcolor{gray!15} \multicolumn{6}{c}{\textit{Zero-shot}} \\
\hline
InternVL3-2B              & 0.269 & 0.070  & InternVL3-2B              & 0.430  & 0.999 \\
SAIL-VL-1.5-2B            & 0.436 & 0.994 & SAIL-VL-1.5-2B            & 0.823 & 0.560  \\
\hline
\end{tabular}
\caption{
Evaluation of the effectiveness and reliability of our ITA and VIR judge models. }
\label{tab:judge-results}
\end{table}

By applying the proposed judge models and filtering criteria, we distill the original 300M captions into a corpus of 250M high-quality image–caption pairs.

\textbf{Chart Caption Data.} 
Beyond general image captions, we curate a large-scale chart caption corpus to improve the model’s chart and table understanding, combining code-based generation with conventional datasets.
We first design a chart data pipeline to automatically and continuously generate high-quality training data. Given a prompt, large language models (LLMs) produce chart code (SVG or Python), along with captions and QA pairs. The code is rendered into chart images, and the images, captions, and QA pairs are stored as standardized triplets for training. The pipeline supports a wide range of chart types, such as bar, line, pie, scatter, area, histogram, heatmap, and box plots, and allows flexible configuration of parameters, including language, model, chart type, and generation rate, while ensuring semantic validity.
To complement this synthetic pipeline, we also collect open-source chart caption datasets in both English and Chinese, including PixMo-cap~\citep{deitke2025molmo} and DVQA~\citep{kafle2018dvqa}, to enhance diversity. Finally, we design an LLM-based annotation workflow for chart captioning and QA generation. Multimodal models annotate both open-source and auto-rendered charts under carefully designed prompts, with manual sampling for quality control. This process yields a large volume of reliable chart annotations. Through this integrated pipeline, we obtain a large corpus of high-quality chart data, consisting of 400K automatically generated captions and 1.29M captions collected from conventional chart datasets, with an equal distribution of English and Chinese.

Through the aforementioned filtering and collection strategies, we construct SAIL-Caption2, which consists of 250M general captions and 1.69M chart captions.

\subsubsection{Synthetic VQA Data} 
Beyond the original caption formulation, we further transform caption data into a QA format using LLMs, thereby scaling up QA data to support efficient and large-scale pre-training of SAIL-VL2. Specifically, we first sample approximately 80M entries from SAIL-Caption2 to construct synthetic VQA data. We then employ a powerful LLM API to convert each caption into multiple question–answer pairs. In this process, the model generates several diverse questions per caption to ensure broad coverage of the visual content, and subsequently produces corresponding answers based on the input text. Although synthetic data may introduce distributional biases in linguistic expression, as LLMs often produce homogenized phrasing with limited variability, we find that the model benefits substantially from such large-scale synthetic data: performance improves smoothly with increasing training budget, exhibiting a logarithmic scaling trend with up to 180M training samples.

\begin{table}[t] 
 \centering 

 \resizebox{0.99\textwidth}{!}{ 
 \begin{tabular}{lcccc} 
 \toprule 
 \textbf{Stages} & \makecell[c]{\textbf{SAILViT} \\ \textbf{Training}} & \makecell[c]{\textbf{Basic} \\ \textbf{Pre-training}} & \makecell[c]{\textbf{Multi-task} \\ \textbf{Pre-training}} & \makecell[c]{\textbf{Supervised} \\ \textbf{Fine-tuning}} \\ 
 \midrule 
 \textbf{Data} & SAIL-Captionv2 & SAIL-Captionv2 & SAIL-Captionv2 & SAIL-Instructionv2 \\ 
  & Caption2QA & OCR & Caption2QA & SAIL-Video \\ 
  & Pure text & & Pure text & \\ 
  & Multimodal VQA & & Multimodal VQA & \\ 
 \midrule 
 \textbf{Tokens} & 121B & 128B & 360B & \makecell[c]{42B -> 40B -> \\ 16B -> 70B} \\ 
 \midrule 
 \textbf{Sequence length} & 8192 & 8192 & 8192 & \makecell[c]{8192 -> 8192 -> \\ 8192 -> 16384} \\ 
 \midrule 
 \textbf{Trainable Params} & \makecell[c]{Connector -> \\ Connector+ViT -> ALL} & Connector & ALL & \makecell[c]{ALL -> ALL -> \\ ALL -> ALL} \\ 
 \bottomrule 
 \end{tabular} 
 } 
 \label{tab:foundational_training} 
  \caption{Overview of foundational pretraining training stages: data composition, token volumes, sequence lengths, and trainable components.} 
 \end{table}

\subsection{Pre-Training Recipe}
We next introduce the pre-training recipe, which consists of two progressive sub-stages designed to gradually activate SAIL-VL2’s multimodal alignment and understanding. Beyond conventional scheduling strategies for LLM/LVM training, we introduce AdaLRS, a dynamic learning rate scheduler applied in the first stage to enable efficient optimization.
\subsubsection{Training Pipeline}
The training pipeline of SAIL-VL2 consists of two progressive stages. The first stage performs basic multimodal pre-training, which equips the model with essential cross-modal alignment and understanding abilities, such as image captioning and OCR. The second stage extends this foundation through multi-task pre-training, enhancing the model’s overall capacity for multimodal understanding and instruction following. 

\textbf{Basic Multimodal Pre-Training.} 
In this stage, we develop SAIL-VL2’s fundamental multimodal alignment and understanding abilities. Starting with a pre-trained SAIL-ViT, a language-pretrained LLM, we train a randomly initialized MLP-based adapter to bridge the gap between vision and language modalities.
A total of 64M samples are used for model training in this stage, including general captions and chart captions from \textit{SAIL-Caption2}, as well as high-quality OCR samples from \textit{IDL-WDS}. 
Instead of training with a fixed learning rate, we set the initial learning rate as $2 \times 10^{-4}$, and apply the AdaLRS algorithm during training, which is elaborated in Section~\ref{sec: adalrs}.
AdaLRS adjusts the initial learning rate upwards to $6.75 \times 10^{-4}$ effectively, surpassing the fixed learning rate baseline by a final loss advantage of over $0.06$. 
The training batch size is set as 2048 in this stage.

\textbf{Multi-task Pre-Training.} 
Following the basic pre-training stage, we further conduct multi-task pre-training to comprehensively strengthen SAIL-VL2’s visual understanding and instruction-following capabilities. 
In this phase, all model parameters are unfrozen for joint optimization. 
Apart from general caption and OCR data as used in the basic pre-training stage, we integrate a series of instruction-tuning datasets for model training, including open-source VQA data, synthetic VQA data, as well as text-based math reasoning samples.
These instruction tuning data not only enhance SAIL-VL2's visual instruction following abilities, but also preserve the LLM's language abilities and therefore enable large-scale vision-language alignment training with visual understanding data. 
In total, 180M samples with equal-sized visual understanding and instruction-tuning data are used in the multi-task pre-training stage.
AdaLRS is not applied in this stage because the training loss on instruction tuning data exhibits poor correlation with model performance. 

\subsubsection{Data Resampling}
To improve training efficiency and data utilization in the pre-training stage of SAIL-VL2, we design a two-step resampling strategy. The goal is to mitigate distributional bias in large-scale caption and VQA data, enhance diversity, and prevent mode collapse, thereby enabling more effective multimodal alignment and instruction tuning.

In the first step, resampling is conducted at the dataset level during the basic pre-training stage. Different datasets naturally exhibit distinct distributional biases. By balancing the sampling ratios across datasets, we increase the diversity of image–text pairs and ensure that the adapter is exposed to heterogeneous visual signals, which supports efficient alignment of vision features into the language space.

In the second step, resampling is performed at the linguistic level during the multi-task pre-training stage. We focus on SAIL-Caption2 and Synthetic VQA datasets, which are large in scale but often show imbalanced language patterns because many samples are generated or post-processed by LLMs. To address this, we rebalance the data at the n-gram level, thereby improving lexical and structural diversity. This linguistic-level resampling boosts data efficiency, allowing SAIL-VL2 to achieve stronger multimodal understanding and more robust instruction-following capabilities

\subsubsection{Adaptive Learning Rate Search}
\label{sec: adalrs}
To achieve more efficient and effective optimization, we introduce an Adaptive Learning Rate Search~\citep{dong2025adalrs} (AdaLRS) strategy during the basic multimodal pre-training of SAIL-VL2. AdaLRS is motivated by an empirical observation that the training loss of LLM/LVM pre-training, as well as the loss descent velocity with respect to the learning rate, exhibits a convex pattern with a shared optimum. Inspired by this finding, AdaLRS dynamically adjusts the learning rate with a backtracking line search strategy based on the loss slope. Specifically: (1) when the loss descent slows down, the learning rate is tentatively increased; (2) if the increase accelerates loss reduction, the adjustment is retained and training proceeds; (3) if the increase further slows convergence, the model parameters and optimizer state are rolled back and the learning rate is decreased instead; (4) through this procedure, suboptimal learning rates can be efficiently adjusted toward the neighborhood of the optimal value within a limited number of steps. The AdaLRS algorithm can be formulated as:
\begin{equation}
\eta_{t+k} = 
\begin{cases} 
\alpha' \eta_t & \text{if } v(\alpha'\eta_t) > v(\eta_t) + 2e \quad (\text{loss slope \textit{increases} $\uparrow$}), \\
\beta' \eta_t & \text{if } v(\alpha'\eta_t) < v(\eta_t) - 2e \quad (\text{loss slope \textit{decreases} $\downarrow$}), \\
\eta_t & \text{otherwise.}
\end{cases}
\label{eq: adalrs}
\end{equation}
$v(\cdot)$ indicates the estimated loss curve slope obtained from a $k$-step window, while $e$ stands for the estimation error between $v$ and the true loss descent velocity $V$.
$\alpha'$ and $\beta'$ are rectified LR scaling factors which satisfy $\alpha'=\max(\lambda^t\alpha, 1), \beta'=\frac{1}{\max(\lambda^t\beta, 1)}$, where $\alpha, \beta > 1$ are co-prime integers and $\lambda \in (0, 1)$ is a decay factor.

\subsubsection{Scaling-law}
\begin{figure*}[h]
\centering
\resizebox{\textwidth}{!}{
\begin{tikzpicture}
\draw (0,0) node[inner sep=0] {\includegraphics[width=\columnwidth, trim={0cm 0cm 0cm 0cm}, clip]{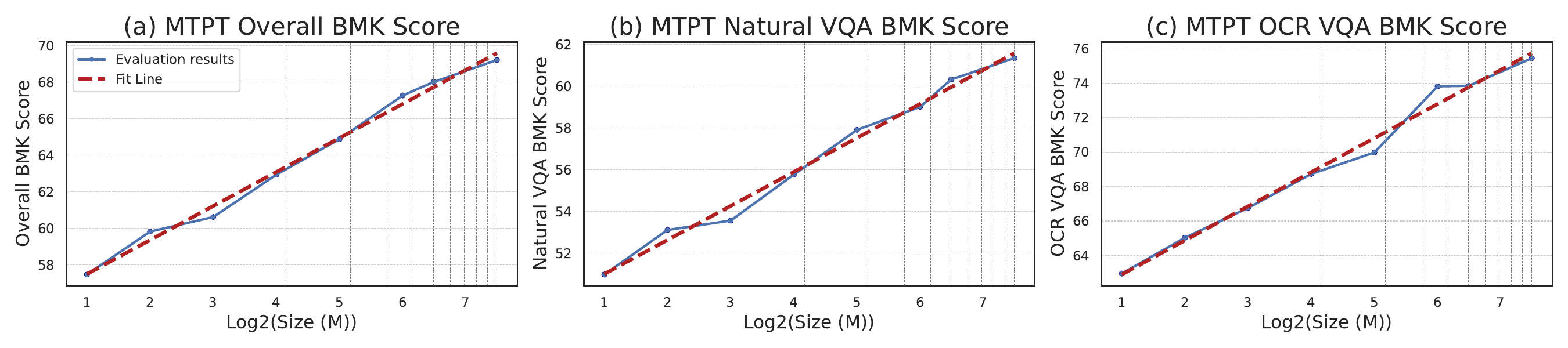}};
\end{tikzpicture}
}
\caption{
    Scaling curves of SAIL-VL2-2B during the multi-task pre-training stage. Results are reported on overall benchmarks, natural-scene VQA datasets, and OCR VQA tasks. 'BMK Score' denotes the average benchmark score.
}
\label{fig: mtpt_scaling}
\end{figure*}

We investigate the impact of scaling multi-task pre-training data on the performance of SAIL-VL2. To this end, we expand the training budget to 360B tokens and systematically evaluate the model across a wide range of benchmark subsets. As illustrated in Figure~\ref{fig: mtpt_scaling}, performance exhibits consistent and monotonic improvements, yielding a smooth empirical scaling curve. The training corpus consists of general captions and VQA data, with approximately $50\%$ of the latter synthesized by annotator models such as SAIL-Captioner~\citep{dong2025scalable} and Qwen3~\citep{qwen3}. While synthetic supervision may introduce linguistic biases, the increased scale and diversity of training data substantially enhance generalization and reasoning. These findings highlight data scaling as a critical factor driving performance gains in the multi-task pre-training stage, and emphasize the importance of large-scale, diverse corpora for advancing multimodal understanding.

\section{Post-Training}

We further enhance SAIL-VL2’s instruction-following ability and incentivize its reasoning capability through a comprehensive post-training stage. The post-training process is structured as a holistic pipeline. It begins with an instruction-tuning SFT stage, aimed at injecting general knowledge and improving fundamental instruction adherence. This is followed by a two-phase thinking-fusion training pipeline that progressively combines SFT with reinforcement learning (SFT–RL) to strengthen the SAIL-VL2’s reasoning ability.

\subsection{Post-Training Data}
In the post-training stage, we organize the data into several categories according to their types and functional roles, and describe the curation process of each in detail.  
 \subsubsection{SAIL-Video}
We initially curate 6.23M video-QA samples from ShareGPTVideo-QA~\citep{zhang2024direct}, NextQA~\citep{xiao2021next}, LongVideoBench-val~\citep{wu2024longvideobench}, PerceptionTest-val~\citep{patraucean2023perception}, LLaVA-Video-QA~\citep{zhang2024video}, and VideoGPT~\citep{yan2021videogpt}. After applying systematic scoring and filtering criteria, we obtain a refined corpus of 5.1M high-quality samples for our SAIL-Video.

For video data filtering and selection, we evaluate quality from two complementary perspectives.
\begin{itemize}
    \item Frame–Instruction Alignment. Most video LVMs~\citep{cheng2024videollama,shu2023audio,wang2024internvideo2,lin2023video,xu2024slowfast} perform tasks such as QA on sampled frames, whereas most existing datasets provide annotations based on continuous video streams. This discrepancy can result in misalignment between sampled frames and instructions, potentially leading to spurious learning or hallucinations.
    \item Data Quality and Task Difficulty. We further assess the overall richness of visual content—capturing element diversity, scene complexity, and information density, as well as the difficulty of associated QA tasks, including reasoning depth, requirements for spatial–temporal understanding, multi-instance interactions, and reliance on external knowledge.
\end{itemize}

To operationalize these criteria, we design specific scoring rules and prompts, defining three metrics: video–QA alignment ($-1$-$10$), video content richness ($-1$-$7$), and QA difficulty ($-1$-$8$). Here, a score of $-1$ indicates that LVM API refused to respond, while higher values represent better quality or greater difficulty. Each video is evaluated under these metrics using the powerful LVM API. For the final selection, we retain videos with alignment and content scores of at least 5 and a difficulty score of at least 3. This procedure yields the final pre-training video corpus.

\subsubsection{SAIL-Instruction2}
Building on the instruction-tuning dataset used in SAIL-VL, SAIL-Instruction, we refine and expand both scale and quality to construct a larger and higher-quality corpus, \textit{SAIL-Instruction2}. Beyond the datasets employed in SAIL-Instruction (\textit{e.g.}, LLaVA-OneVision\citep{li2024llava} and Cauldron\citep{laurenccon2024matters}), we incorporate additional high-quality open-source instruction-tuning datasets, including Mammoth\citep{guo2024mammoth}, MMPR\citep{wang2024enhancing}, and Molmo~\citep{deitke2024molmo}, thereby providing broader coverage and stronger supervision for instruction-following pre-training. To enhance effectiveness, we sample long-answer and reasoning-oriented instances from the LLaVA-CoT, MMPR, and Condor datasets, fostering more complex reasoning behaviors. Following SAIL-VL~\citep{dong2025scalable}, all incremental data undergo a two-stage validation process: (i) a quality evaluation to ensure intrinsic reliability, and (ii) an incremental evaluation to verify performance gains when integrated with existing data. Only data passing both stages is retained.

To enhance diversity and mitigate distributional bias, we employ a latent-class-based filtering strategy. For each VQA sample, an LVM generates a descriptive phrase, from which logits and latent embeddings are extracted to define its latent class. This approach maps heterogeneous samples into a unified semantic space, effectively bridging the representation gap across datasets. We enhance the original coarse-level dataset classification by clustering latent classes into fine-grained semantic buckets, thereby expanding the category space by nearly tenfold. Using these refined classification buckets, we perform uniform sampling and re-annotate the samples with APIs from state-of-the-art closed-source models, improving both dataset accuracy and consistency. This strategy not only balances the sample distribution but also enhances overall data quality. Finally, 20M high-quality, diverse samples form our SAIL-Instruction2 dataset.

\begin{figure}[h]
    \centering
    
    \includegraphics[width=0.65\linewidth]{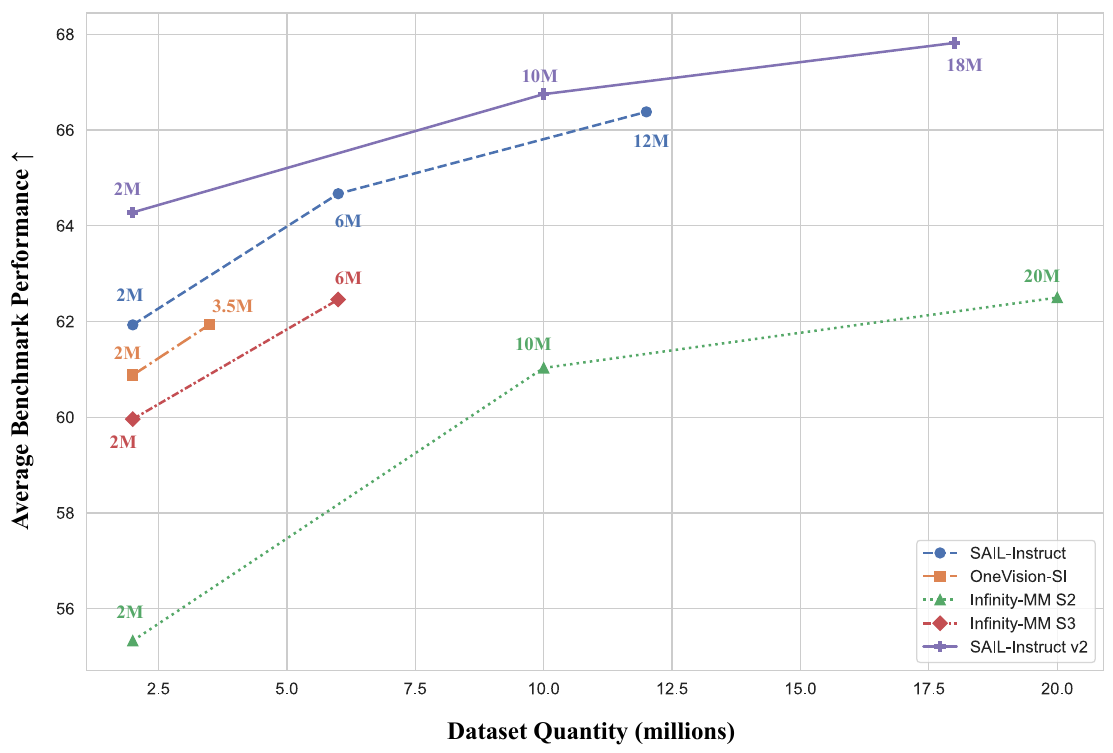}
    \caption{Analysis of instruction data quality and scale on model performance. Using SAIL-VL2-2B as the base architecture after multi-task pre-training, we compare SFT results across different datasets and data scales.}
    \label{fig:data-analysis}
\end{figure}

\textbf{Instruction Data Analysis.} To investigate the effect of instruction data on model performance, we conduct a systematic comparison using SAIL-VL2-2B as the base architecture after multi-task pre-training. As illustrated in Figure~\ref{fig:data-analysis}, we evaluate SFT results across different datasets and data scales. Compared with prior instruction datasets, our SAIL-Instruction2 consistently delivers superior performance under the same data budget. This demonstrates that SAIL-Instruction2 provides higher-quality supervision and validates the effectiveness of our data optimization and filtering pipeline. These findings underscore the critical role of curated and diversified instruction data in enhancing multimodal instruction-following capabilities.

\subsubsection{Multimodal CoT Data}  
To cultivate strong multimodal reasoning capabilities, we construct a large-scale dataset by aggregating diverse public sources and organizing them into two complementary subsets for the LongCoT SFT, Think-Fusion SFT and the subsequent Reinforcement Learning. The SFT corpus integrates established visual-instruction and mathematical reasoning datasets, including VisualWebInstruct~\citep{jia2025visualwebinstruct}, MathV360K~\citep{shi2024math}, and LLaVA-CoT~\citep{xu2024llava}, providing the model with structured reasoning supervision. For the RL stage, we expand the scope to cover a broader range of tasks—Math~\citep{sun2024mm,meng2025mm,xiao2025fast,zhang2025cmmcot}, Puzzle~\citep{chia2024puzzlevqa}, Science~\citep{wang2025vl, lu2022learn}, OCR~\citep{chen2025learning}, and Counting~\citep{johnson2017clevr}—to enhance generalization and robustness. Since these heterogeneous sources inevitably contain noise and formatting inconsistencies, we employ a rigorous data curation pipeline, described in the following section. After refinement, the final dataset comprises 400K samples for LongCoT SFT, 1M samples for Think-Fusion SFT, 50K samples for RL with verifiable rewards, and 100K samples for RL with a mixed reward system.

\begin{table}[h!] 
 \centering 

 \resizebox{0.99\textwidth}{!}{ 
 \begin{tabular}{lcccc} 
 \toprule 
 \textbf{Stages} & \makecell[c]{\textbf{LongCoT} \\ \textbf{SFT}} & \makecell[c]{\textbf{Verifiable-Reward} \\ \textbf{RL}} & \makecell[c]{\textbf{Think-Fusion} \\ \textbf{SFT}} & \makecell[c]{\textbf{Mixed-Reward} \\ \textbf{RL}} \\ 
 \midrule 
 \textbf{Data} & Multimodal CoT Data & STEM RL data & Multimodal CoT Data & STEM RL Data \\ 
  & & & SAIL-Instructionv2 & General RL Data \\ 
 \midrule 
 \textbf{Tokens} & 0.4B & 0.1B & 1B & 0.2B \\ 
 \midrule 
 \textbf{Sequence length} & 20480 & 20480 & 20480 & 20480 \\ 
 \midrule 
 \textbf{Trainable Params} & ALL & ALL & ALL & ALL \\ 
 \bottomrule 
 \end{tabular} 
 } 
 \label{tab:advanced_training_stages} 
  \caption{Overview of advanced post-training stages: data composition, token volumes, sequence lengths,
and trainable components stages.} 
 \end{table}

\subsection{Post-Training Recipe}
The post-training pipeline consists of three stages. The first stage focuses on foundational SFT to develop basic multimodal understanding, followed by a reasoning-oriented tuning pipeline. This pipeline includes a LongCoT SFT, a verifiable reward-driven RL, a Thinking-Fusion SFT, and a mixed reward-driven RL tuning phase.

\subsubsection{Basic Supervised Fine-Tuning}

The Basic SFT stage is organized as a progressive knowledge injection pipeline. The first three phases focus on training the image modality, employing a curriculum learning strategy to gradually introduce increasingly complex data as training progresses, thereby fine-tuning all parameters of SAIL-VL2.

In the first phase, SAIL-VL2 learns foundational instruction-following abilities while incorporating world knowledge from the Infinity-MM Stage2 dataset~\citep{gu2024infinity}. The second phase optimizes the model’s visual instruction-following capabilities using SAIL-Instruction2, a newly curated 20M-sample high-quality visual instruction dataset. In the third phase, the model is exposed to a challenging subset of visual instruction data, including LLaVA-CoT, MMPR, and Condor datasets, enabling it to tackle a wider range of complex instruction-following tasks. To further enhance SAIL-VL2’s understanding of video while preserving its proficiency in image comprehension, we introduce a video-image mixture training strategy in the final phase. This strategy combines high-quality image data from SAIL-Instruction2 with carefully filtered video data from the SAIL-Video dataset, maintaining a balanced 1:1 ratio between video and image data, thus ensuring effective multimodal learning.

\noindent\textbf{Model Soup.} To further enhance model performance after the SFT, we adopt a simple yet effective model soup strategy. We first categorize trained models into two groups: homogeneous models, which follow similar optimization trajectories such as comparable hyperparameter settings and data compositions, and heterogeneous models, which differ in these aspects. As shown in Table~\ref{tab:model-merging}, experimental results reveal that merging homogeneous models consistently yields stable and notable performance improvements, whereas merging heterogeneous models often causes performance degradation. Based on these observations, our model soup strategy focuses on merging homogeneous models to achieve reliable performance gains.
\begin{table}[t]
\centering
\renewcommand\arraystretch{1.3}
\resizebox{\textwidth}{!}{%
\setlength{\tabcolsep}{2pt} 
\begin{tabular}{l|c|ccccccccccc}
\hline
\textbf{Model} & \textbf{AVG} & \textbf{ChartQA} & \textbf{DocVQA} & \textbf{InfoVQA} & \textbf{MMBench$_{\text{v1.1}}$} 
& \textbf{MMMU} & \textbf{MMStar} & \textbf{OCRBench} & \textbf{TextVQA} & \textbf{RealWorldQA} \\
\hline
\rowcolor{gray!15} \multicolumn{11}{c}{\textit{Homologous Model Merging}} \\
\hline
Base model 1 & \underline{74.91} & \underline{83.88} & \underline{91.98} & \underline{74.03} & \underline{80.15} & \underline{45.89} & \underline{62.60} & \underline{87.00} & 80.91 & 67.71 \\
Base model 2 & 74.54 & 82.80  & 91.24 & 72.44 & 79.33 & 45.78 & 61.27 & 87.00 & \underline{81.06} & \underline{69.97} \\
\midrule
Merge model  & \textbf{76.60}  & \textbf{85.60}  & \textbf{92.22} & \textbf{75.35} & \textbf{81.46} & \textbf{48.56} & \textbf{63.33} & \textbf{89.50} & \textbf{82.05} & \textbf{71.37} \\
\hline
\rowcolor{gray!15} \multicolumn{11}{c}{\textit{Heterologous Model Merging}} \\ 
\hline
Base model 1 & \underline{74.54} & \underline{82.80}  & \underline{91.24} & \underline{72.44} & \underline{79.33} & \textbf{45.78} & \underline{61.27} & \underline{87.00} & \underline{81.06} & \textbf{69.97} \\
Base model 2 & \textbf{75.41} & \textbf{84.56} & \textbf{92.30}  & \textbf{74.71} & \textbf{80.19} & \underline{45.11} & \textbf{63.27} & \textbf{88.40} & \textbf{81.66} & \underline{68.50}  \\
\midrule
Merge Model  & 12.86 & 10.36 & 5.25  & 10.47 & 12.73 & 37.22 & 30.73 & 0.12 & 8.83  & 47.32 \\
\hline
\end{tabular}
}
\caption{Performance of Homologous and Heterologous Model Merging on Various Datasets}
\label{tab:model-merging}
\end{table}

\subsubsection{LongCoT Supervised Fine-tuning}

The first stage of our post-training framework is LongCoT Supervised Fine-tuning, designed specifically to enhance the model's step-by-step reasoning capabilities for complex problems. This is achieved by fine-tuning the model exclusively on a large-scale, high-quality Long-Chain-of-Thought (LongCoT) dataset.

\noindent\textbf{LongCoT Data Curation.}
We develop a holistic data pipeline to build our high-quality LongCoT dataset. We first collect diverse datasets, including VisualWebInstruct~\citep{jia2025visualwebinstruct}, MathV360K~\citep{shi2024math}, and LLaVA-CoT~\citep{xu2024llava}, and then apply a cleaning protocol which includes: (i) removing content extraneous to the reasoning process, such as system prompts and conflicting hints; (ii) unifying the output format by placing the thinking process within <think> and </think> tags, and the final answer in the  \verb|\boxed{}| tags; and (iii) deduplicating samples based on image and question pairs. For CoT data generation, we employ a guided prompting strategy where we provide both the question and its ground-truth, tasking the model to generate a detailed reasoning process that logically connects the two. Following this generation, we execute a rigorous filtering workflow. This workflow includes a final deduplication pass, format validation, and three critical quality filters: (i) Redundancy Filtering: We measure the token overlap between the generated CoT and the final answer, discarding samples with high similarity to penalize trivial reasoning. (ii) Answer Distillation: We utilize a judge model to refine and shorten overly verbose ground-truth answers, mitigating the tendency for models to overthink. (iii) CoT Length Balancing: We analyze the token length distribution of the generated CoT and resample to ensure a balanced representation of reasoning complexities. We finally get 400K LongCoT samples.

\noindent\textbf{Training Recipe.}
We fine-tune the model on the 400K LongCoT samples. All model parameters are trained for one epoch using the AdamW optimizer with a global batch size of 1024. We employ a cosine learning rate schedule with a peak learning rate of $1 \times 10^{-6}$. The optimization objective is a standard next-token prediction over the entire output sequence:

\begin{equation}
\label{eq:longcot_sft_loss}
\mathcal{L}_{\text{LongCoT SFT}} = - \frac{1}{|D_{\text{CoT}}|} \sum_{(I, T, A) \in D_{\text{CoT}}} \log P_\theta(T \circ A | I)
\end{equation}

where $D_{\text{CoT}}$ is the LongCoT dataset, $I$ is the input instruction, $T$ is the thinking process, $A$ is the final answer, and $\circ$ denotes string concatenation. This training strategy explicitly teaches the model to generate a detailed reasoning process before providing the final answer, thereby significantly enhancing its ability to solve complex problems that require step-by-step thinking.

\subsubsection{Reinforcement Learning with Verifiable Rewards}

Following the LongCoT Supervised Fine-tuning stage, we further enhance the model's reasoning capabilities through a Reasoning Reinforcement Learning (RL) stage. This stage refines the model by optimizing it against a reward system focused on two primary objectives: the correctness of the final answer and adherence to the specified output format.

\noindent\textbf{Data Curation.}
The dataset for our RL stage is constructed from a diverse collection of public sources, covering tasks such as Math~\citep{sun2024mm, meng2025mm}, Puzzle~\citep{chia2024puzzlevqa}, Science~\citep{wang2025vl,lu2022learn}, OCR~\citep{chen2025learning}, and Counting~\citep{johnson2017clevr}. This raw data is then refined through a two-stage filtering pipeline. The first stage mitigates reward hacking by employing an LLM to convert multiple-choice questions with numerical or symbolic answers into a free-response format. The second stage applies a difficulty-based filtering strategy using the model from the previous stage, evaluated on its \texttt{pass@4} score. Specifically, for tasks that do not require explicit reasoning, we discard trivial problems where the model achieves a perfect score (\texttt{pass@4}=1).  For reasoning-heavy tasks, we apply a more stringent filter, retaining problems that are challenging yet solvable by removing both the easiest (\texttt{pass@4}=1) and hardest (\texttt{pass@4}=0) samples. This pipeline yields a dataset of challenging yet tractable problems for stable RL training. We finally got 70K stem samples for our RL training.

\noindent\textbf{Reward System.}
Our reward system is designed to guide the model towards generating accurate and well-structured responses. It is composed of two main signals: (1) Answer Reward: This is a rule-based reward that checks for the correctness of the final answer contained within the \verb|\boxed{}| tag. For problems with verifiable solutions, such as mathematical calculations, it provides a clear, objective signal of success; (2) Format Reward: This reward ensures the model's output adheres to the required structure, specifically by verifying that the reasoning process is correctly enclosed within <think> and </think> tags. Both reward signals are discrete binary values (0 or 1), providing direct and unambiguous feedback for RL training.

\noindent\textbf{Training Recipe.}
We optimize the LongCoT SFT model using a Proximal Policy Optimization (PPO)-based~\citep{schulman2017proximal} algorithm to learn a single, unified reasoning policy. To accommodate different model architectures, we employ specialized PPO variants. For our dense model, we use DAPO~\citep{yu2025dapo}, a memory-efficient optimizer that enhances training stability. For our Mixture-of-Experts (MoE) model, we use GSPO~\citep{zheng2025group}, which provides stable and targeted updates to individual experts. The training is configured with a context length of 16384 and a maximum generation length of 4096 tokens. In each training episode, we generate 2048 rollouts from the policy and perform 8 gradient updates using a mini-batch size of 512. The learning rate for the policy network is set to $1 \times 10^{-6}$. To encourage exploration, the PPO clipping value is dynamically adjusted within the range of [0.20, 0.28].

\subsubsection{Think-Fusion Supervised Fine-tuning}

Following the initial LongCoT SFT and RL with Verified Reward, the next stage of our post-training framework is \textbf{Think-Fusion Supervised Fine-tuning}. It is designed to simultaneously enhance the model's reasoning capabilities while maintaining its broad general understanding. This is achieved by fine-tuning the model on a mixed dataset strategically composed to address this dual objective.

\noindent\textbf{Data Composition and Sampling.}
The dataset for this stage is constructed as a strategic mixture to balance two key objectives: maintaining the model's general capabilities while enhancing its specialized reasoning skills. The large majority, approximately 90\% of the data, consists of general-purpose, direct-answer instruction pairs. This component covers a wide range of common tasks and serves to reinforce the model's wide-ranging abilities, ensuring it remains a capable and versatile assistant. The remaining 10\% of the dataset is composed of high-quality Chain-of-Thought (CoT) exemplars. This smaller, focused portion is specifically chosen to cultivate the model's ability to perform complex, step-by-step reasoning. Crucially, these are not just any randomly selected CoT examples. Instead, they are carefully harvested from our preceding reinforcement learning (RL) stage through a process of rejection sampling. This means we selected only the best-performing reasoning chains that the model itself generated, ensuring the examples are both correct and logically sound. In the end, we get a final dataset for this stage comprising 100K high-quality CoT samples and 900K direct question-answering samples, creating a total of one million training instances.

\noindent\textbf{Training Recipe.} We fine-tune the model on the mixed dataset, and all model parameters are trained for one epoch using the AdamW optimizer with a global batch size of 1024. We employ a cosine learning rate schedule with a peak learning rate of $1 \times 10^{-6}$. The optimization objective is a standard next-token prediction, where the loss is applied conditionally to teach the model both response styles. For direct-answer pairs $(I, A)$, where $I$ is the instruction and $A$ is the answer, the loss is computed only on the answer tokens. For CoT exemplars $(I, T, A)$, where $T$ is the thinking process, the loss is computed across the entire sequence, including both the thinking process and the final answer. This dual-target loss function can be formally denoted as:

\begin{equation}
\small
\label{eq:sft_loss}
\mathcal{L}_{\text{Think-Fusion SFT}} = - \frac{1}{|D|} \left( \sum_{(I, T, A) \in D_{\text{CoT}}} \log P_\theta(T \circ A | I) + \sum_{(I, A) \in D_{\text{direct}}} \log P_\theta(A | I) \right)
\end{equation}

where $D$ is the entire mixed dataset, and $\circ$ denotes string concatenation. This training strategy explicitly teaches the model to adapt its output, providing concise answers for simple queries while generating detailed reasoning for complex problems that require it.

\subsubsection{Reinforcement Learning with a Mixed Reward System}

Following the Think-Fusion Supervised Fine-tuning stage, we further enhance the model's reasoning capabilities through an RL stage. This stage is guided by a mixed reward system that provides comprehensive feedback beyond a single correctness score. Our reward system integrates three primary signals: the Answer Reward, which assesses final answer quality; the Thinking Reward, which evaluates the reasoning process; and a Format Reward, which encourages adherence to the desired output structure. Notably, for tasks where ground truth is not deterministically verifiable, we leverage an LVM-based judge to provide a nuanced reward signal. By optimizing the model with this comprehensive reward system on a diverse dataset, we enhance its performance on multimodal reasoning tasks, ensuring both high accuracy and logical coherence.

\noindent\textbf{Data Curation.}
To further challenge the model, we refine the dataset from the preceding Reasoning RL stage. We apply a rejection sampling methodology to the previous set of reasoning-intensive samples, a process that filters for "hard cases" where the model's performance was suboptimal. This creates a more challenging, curated dataset of 50K stem samples. To ensure the model retains its broad conversational and instruction-following capabilities while improving on these hard cases, we combine these specialized samples with 50K new general-purpose samples from LLaVA-OneVision~\citep{li2024llava}. The final training dataset, therefore, consists of 100K samples, balancing advanced reasoning with general utility.

\noindent\textbf{Reward System.}
Our mixed reward system is comprised of three distinct signals, designed to guide the model towards robust and trustworthy reasoning. The first is an \textit{Answer Reward} that scores the final answer, using a hybrid of rule-based checks for verifiable problems and an LVM-based judge for nuanced tasks. The second is a \textit{Think Reward} that holistically evaluates the entire reasoning path on its logical soundness, factual grounding, and answer consistency. The third is a \textit{Format Reward} to ensure adherence to a specific output structure, with the thinking process to be in <think> and </think> tags and the final answer in a \verb|\boxed{}| tag. All reward signals are discrete binary values (0 or 1), and they are integrated into a final reward formulation where the coefficients of the three components sum to 1.

\noindent\textbf{Training Recipe.}
We follow the same training recipe as the RL with Verifiable Rewards stage described previously. The Think-Fusion SFT model is optimized using PPO-based algorithms (DAPO for dense models, GSPO for MoE models) on the curated 150K sample dataset. This optimization process is carried out on our curated dataset of 150K high-quality samples. To maintain a controlled experimental setup, all training hyperparameters—such as learning rate, batch size, and optimizer settings—are kept identical to the prior stage. A key outcome we observe from this process is that the model effectively internalizes the reasoning capabilities it was taught. In other words, it learns the logical pathways to the correct answer without needing to explicitly write out each step. This allows it to provide accurate direct answers even when a user does not specifically prompt it for a step-by-step thinking process, making the final model both powerful and efficient in practice.

\section{Training Infrastructure}
To ensure efficiency and robustness during training, we design a systematic infrastructure framework tailored for large-scale multimodal models.
\subsection{Stream Packing}
To enhance training efficiency and reduce computational waste, we design a systematic stream packing strategy that jointly optimizes the handling of language and vision tokens. This strategy integrates two complementary components: batching with online packing and visual packing, ensuring balanced workloads, reduced memory footprint, and even improved model performance.

\noindent\textbf{Batching and Online Packing.} In conventional LVM training, all sequences in a batch are padded to the length of the longest sequence, resulting in severe inefficiency when sample lengths are highly skewed. We address this by concatenating variable-length samples into continuous streams, thereby minimizing the number of padding tokens. To preserve independence across samples, we adjust positional embeddings and attention masks accordingly. To further improve efficiency, each node maintains a buffer of candidate samples, from which micro-batches are dynamically constructed. This online packing procedure enhances sample diversity, maximizes GPU utilization by filling sequences to the hardware-supported maximum length, and achieves balanced workloads across devices. Moreover, to prevent starvation of long samples that exceed the maximum sequence length, we enforce their periodic inclusion during batch construction.

\noindent\textbf{Visual Packing.} Beyond balancing text token lengths, effective LVM training must also address discrepancies in visual token counts. In practice, different samples may produce substantially different numbers of visual tokens, particularly in models such as SAIL-VL2-AnyRes, where original image resolutions are preserved, leading to workload imbalance across devices and increased computational cost for the vision encoder. To mitigate this, we extend stream packing with a visual token balancing constraint. During sample selection from the buffer, we impose an additional condition that equalizes the number of visual tokens across devices. This ensures balanced computational loads for both the LLM and the vision encoder, thereby improving overall training efficiency.

\noindent\textbf{Training Efficiency and Performance Analysis.}
Our packing strategies bring significant improvements in both training efficiency and model performance. Compared with the baseline, data packing nearly doubles Streaming Multiprocessor (SM) utilization and accelerates training by $50\%$, while visual packing alleviates excessive memory usage in the vision encoder, leading to a further $48\%$ gain in efficiency. Beyond efficiency, ablation studies show that models trained with packing achieve an average $+0.7\%$ improvement over the baseline, with particularly strong gains on open-ended QA benchmarks such as LLaVA-Bench and MMVet, owing to the mitigation of sequence truncation and more effective training on long-context multimodal inputs.

\subsection{MoE Infra}
Training Mixture of Experts (MoE) models presents substantial challenges arising from their massive parameter scale and the inherent imbalance in expert activation. Although MoE architectures reduce per-step computation compared to dense models, the overall training process remains dominated by two sources of overhead: the computational cost of activating multiple experts and the communication burden associated with intensive inter-device data exchange. To alleviate computational overhead, we implement optimized kernel fusion for expert operations, achieving up to a 3× training speedup, and extend these optimizations to core modules such as self-attention and Layer Normalization, thereby improving end-to-end efficiency. To address communication inefficiencies, we design hardware-adapted infrastructures: on NPU processors, we employ a Megatron-based distributed framework that partitions both MoE and ViT across devices, combining pipeline and expert parallelism to reduce memory footprint and communication overhead through efficient weight sharing; on NVIDIA devices (\textit{e.g.}, H20, A100), we adopt DeepSpeed ZeRO-2~\citep{lan2025zenflow} with CPU offloading, which provides superior communication efficiency relative to ZeRO-3 and directly mitigates the computation–communication imbalance intrinsic to MoE training.

\section{Experiment}
In this section, we present a comprehensive experimental comparison and analysis across multiple dimensions using a wide range of LVM benchmarks
\subsection{Evaluation Setting}
We first outline the detailed evaluation settings used in our experimental analysis. 

\noindent\textbf{Model Zoo.} We introduce the model zoo for SAIL-VL2, encompassing a range of model scales and configurations. Equipped with our SAIL-ViT model family, SAIL-VL2 employs foundational LLM components such as Qwen3-0.6B, Qwen3-1.7B, Qwen3-8B, Qwen3-16B-A2.5B, and Qwen3-30B-A3B for comprehensive pre-training. Based on the various settings of SAIL-ViT, we detail the configurations of the SAIL-VL2 zoo as follows:
\begin{itemize}
    \item \textit{Fixed-Resolution SAIL-ViT:} In this setting, we preprocess the images by cropping and resizing them to a base resolution of 448×448. A 2×2 pixel shuffle downsampling strategy is applied to each frame to improve computational efficiency. Based on this configuration, we trained and provided the SAIL-VL2-1B, SAIL-VL2-2B, SAIL-VL2-8B, and SAIL-VL2-30B-A3B models.
    \item  \textit{SAIL-ViT-AnyRes:} This setting supports input images of arbitrary resolutions, preserving original aspect ratios while enabling more fine-grained modeling. The largest supported resolution, given a maximum input length of 16,384, is 1792×1792. Using this approach, we trained the SAIL-VL2-AnyRes-2B model.
\end{itemize}

Furthermore, we extend the SAIL-VL2 models with a thinking version via thinking-fusion tuning (SFT-RL combined tuning). These models, marked as SAIL-VL2-2B-think, SAIL-VL2-8B-think, and SAIL-VL2-30B-A3B-think, provide enhanced reasoning capabilities.

\noindent\textbf{Baselines.} To conduct a comprehensive performance analysis of SAIL-VL2, we present an extensive set of baselines for comparison. In the basic setting, SAIL-VL2 is compared with leading open-source and proprietary models, including, but not limited to, the Qwen2.5-VL~\citep{qwen2_5_vl}, InternVL3~\citep{internvl3}, InternVL3.5~\citep{wang2025internvl3}, Ovis2~\citep{lu2025ovis2}, and Ovis-U1~\citep{wang2025ovis}. Detailed model performance comparisons are provided in Table~\ref{tab:base-4b}, Table~\ref{tab:base-8b} and Table~\ref{reasoning_vl}.

For the thinking model, we compare SAIL-VL2 with both open-source and proprietary models, such as Kimi-VL-A3B-Thinking-2506, Keye-VL-8B-Thinking OpenVLThinker-7B, and WeThink-7B. Further detailed comparisons of model performance are shown in Table~\ref{reasoning_vl}.

\noindent\textbf{Benchmarks.}
We perform extensive experimental comparisons and analysis across a diverse set of benchmarks, covering 106 datasets. These benchmarks are organized into four primary evaluation dimensions: General Open-source Multimodal Understanding Benchmarks, the OpenCompass evaluation system~\citep{2023opencompass}, and open-source video understanding benchmarks:
\begin{itemize}
    \item The General Open-source Multimodal Understanding Benchmarks encompass 72 datasets, spanning a wide range of tasks including "Multi-Image Understanding," "Multilingual Understanding," "Real-World Understanding," "Charts, Documents, and OCR Tasks," "Multimodal Reasoning and Mathematics," "Comprehensive Multimodal Assessment," "Visual Localization," "Multimodal Hallucination Assessment," and "Instruction Following."
    \item The OpenCompass dimension includes 8 datasets, aligned with the official OpenCompass leaderboard, covering tasks from multimodal understanding to reasoning.
    \item The open-source video understanding benchmarks consist of 9 datasets, incorporating all 5 video evaluation sets from the official OpenCompass video leaderboard, as well as four additional benchmarks: ActivityNetQA, LongvideoBench, NextQA, and TVBench.
\end{itemize}

Based on the comprehensive benchmarks mentioned above, we conduct a thorough performance evaluation and comparison with current methods, providing a holistic analysis of SAIL-VL2's capabilities.

\noindent\textbf{Evaluation Protocol.}
For the comparison of SAIL-VL2 basic models, we adopt a unified evaluation framework based on a customized version of VLMEvalKit~\citep{2023opencompass}. For datasets requiring LLM-assisted assessment, we employ Doubao-1.5-vision-pro-32k-250115~\citep{seed_1_5_vl} as the evaluation API, and all baseline models are re-evaluated under this setting. All evaluation results for the SAIL-VL2-Thinking models are sourced from the official OpenCompass open-source leaderboard, with the exception of SAIL-VL2-A3B-Thinking and Keye-VL-8B-Thinking. To ensure a fair comparison for these two models, we conducted the evaluation using a customized VLMEvalKit with GPT-4O-Mini as the judge model. This process strictly aligns with the official OpenCompass settings, making our results directly comparable to the leaderboard.

\subsection{Main Results}
Based on the aforementioned settings, we conduct a comprehensive performance evaluation of SAIL-VL2 across these benchmarks. First, we analyze the performance of SAIL-ViT in detail. We then evaluate all basic (non-thinking) models in the SAIL-VL2 series using this extensive benchmark, presenting the results across several widely used evaluation datasets in Table~\ref{tab:base-4b} and Table~\ref {tab:base-8b}. The 'thinking' models of SAIL-VL2 were assessed and compared exclusively on the official OpenCompass Reasoning leaderboard, as shown in Table~\ref {reasoning_vl}.  Experimental results show that SAIL-VL2-2B/8B and SAIL-VL2-Thinking-8B achieve state-of-the-art average performance on the OpenCompass evaluation, consistently surpassing all publicly available open-source models at comparable parameter scales.

\subsubsection{Experimental Analysis on SAIL-ViT}
We verify the effectiveness of SAIL-ViT through two complementary evaluations: (1) its core visual classification performance, and (2) its capability to align visual features with multimodal representations.

\noindent\textbf{Zero-shot Image Classification of SAIL-ViT.}
To evaluate the multimodal representation capabilities of SAIL-ViT, we conducted a comparative analysis with open-source ViT models of similar scale, focusing on the purely visual task of image classification. Specifically, we averaged the token outputs from the final block of various visual backbones, which were then processed through a two-layer MLP to generate a token representing the probability value. All models were trained for 10 epochs using the ImageNet-1k training dataset, followed by evaluation on four test datasets: ImageNet-1k~\citep{russakovsky2015imagenet}, ImageNet-A~\citep{hendrycks2019nae}, ImageNet-R~\citep{hendrycks2021many}, and ImageNet-v2~\citep{recht2019imagenet}. The experimental results are presented in Table~\ref{tab:vit-image-clas}. Under consistent experimental settings, SAIL-ViT-Large demonstrated significant improvements over the baseline model AIMv2~\citep{aimv2} on a CV benchmark consisting of four test sets, with an average gain of $+1.5\%$. For SAIL-ViT-Huge, the average improvement was $+2.11\%$. Additionally, by increasing the data scale during the world knowledge infusion stage, we developed SAIL-ViT, which further enhanced vision task performance by $+2.73\%$ compared to the baseline. Overall, these results highlight the superior performance and scalability of SAIL-ViT compared to baseline ViT models on image classification benchmarks.
\begin{table}[t]
\centering

\setlength{\tabcolsep}{2pt}
\resizebox{\textwidth}{!}{%
\begin{tabular}{l|ccccccc}
\toprule
\textbf{Settings}    & \multicolumn{1}{l}{\textbf{AIMv2-Large}} & \begin{tabular}[c]{@{}c@{}}\textbf{InternViT-300M-}\\ \textbf{448px-V2.5}\end{tabular} & \multicolumn{1}{l}{\textbf{SAIL-ViT-Large}} & \multicolumn{1}{l}{\textbf{AIMv2-Huge}} & \multicolumn{1}{l}{\textbf{SAIL-ViT-Huge}} & \multicolumn{1}{l}{\textbf{SAIL-ViT-Huge-v2}} & \begin{tabular}[c]{@{}c@{}}\textbf{InternViT-6B-}\\ \textbf{448px-V2.5}\end{tabular} \\ \midrule
ImageNet-1k & 79.88 & 73.70  & 80.71  & 81.68 & 82.21 & 82.34 & 84.18 \\
ImageNet-A  & 25.41 & 13.45  & 29.31  & 28.96 & 33.04 & 35.63 & 46.27 \\
ImageNet-R  & 55.73 & 39.99 & 56.42  & 58.13 & 60.33 & 59.90 & 59.88 \\
ImageNet-V2 & 76.45 & 69.55  & 77.06  & 77.32 & 78.94 & 79.12 & 80.92 \\
Average     & 59.37 & 49.17  & 60.87  & 61.52 & 63.63 & 64.25 & 67.81 \\ \bottomrule
\end{tabular}
}
\caption{Comparison results of different visual backbones on visual recognition tasks.}
\label{tab:vit-image-clas}
\end{table}

\begin{table}[h]
\centering
\setlength{\tabcolsep}{6pt}
\begin{tabular}{ll|ccc}
\toprule
\multicolumn{1}{l}{\textbf{LLM}} &
\multicolumn{1}{l|}{\textbf{ViT}} &
\begin{tabular}[c]{@{}c@{}}\textbf{Average nearest neighbor} \\
\textbf{distance}\end{tabular} 
& \begin{tabular}[c]{@{}c@{}}\textbf{Wasserstein} \\ \textbf{distance}\end{tabular} & 
\begin{tabular}[c]{@{}c@{}}\textbf{Mean} \textbf{overall} \\ \textbf{distance} \end{tabular} 
\\
\midrule
Qwen3-0.6B         & Aimv2      & 1.42 & 4.86 & 10.78 \\
Qwen3-0.6B         & SAIL-ViT   & 1.15 & 3.88 & 9.52  \\
\midrule
Qwen3-1.7B         & Aimv2      & 1.30 & 4.65 & 11.11 \\
Qwen3-1.7B         & SAIL-ViT   & 1.05 & 3.60 & 9.89  \\
\midrule
Qwen3-8B           & Aimv2      & 0.78 & 3.59 & 11.24 \\
Qwen3-8B           & SAIL-ViT   & 0.66 & 2.63 & 10.06  \\
\midrule
Interlm2.5-1.8B    & Aimv2      & 1.17 & 4.14 & 10.99 \\
Interlm2.5-1.8B    & SAIL-ViT   & 0.96 & 3.18 & 9.79  \\
\bottomrule
\end{tabular}
\caption{Comparison of LLM and ViT models on three distance metrics.}
\label{tab:llm-vit-distances}
\end{table}

\begin{figure*}[t]
\centering
\resizebox{\textwidth}{!}{
\begin{tikzpicture}
\draw (0,0) node[inner sep=0] {\includegraphics[width=\columnwidth, trim={0cm 0cm 0cm 0cm}, clip]{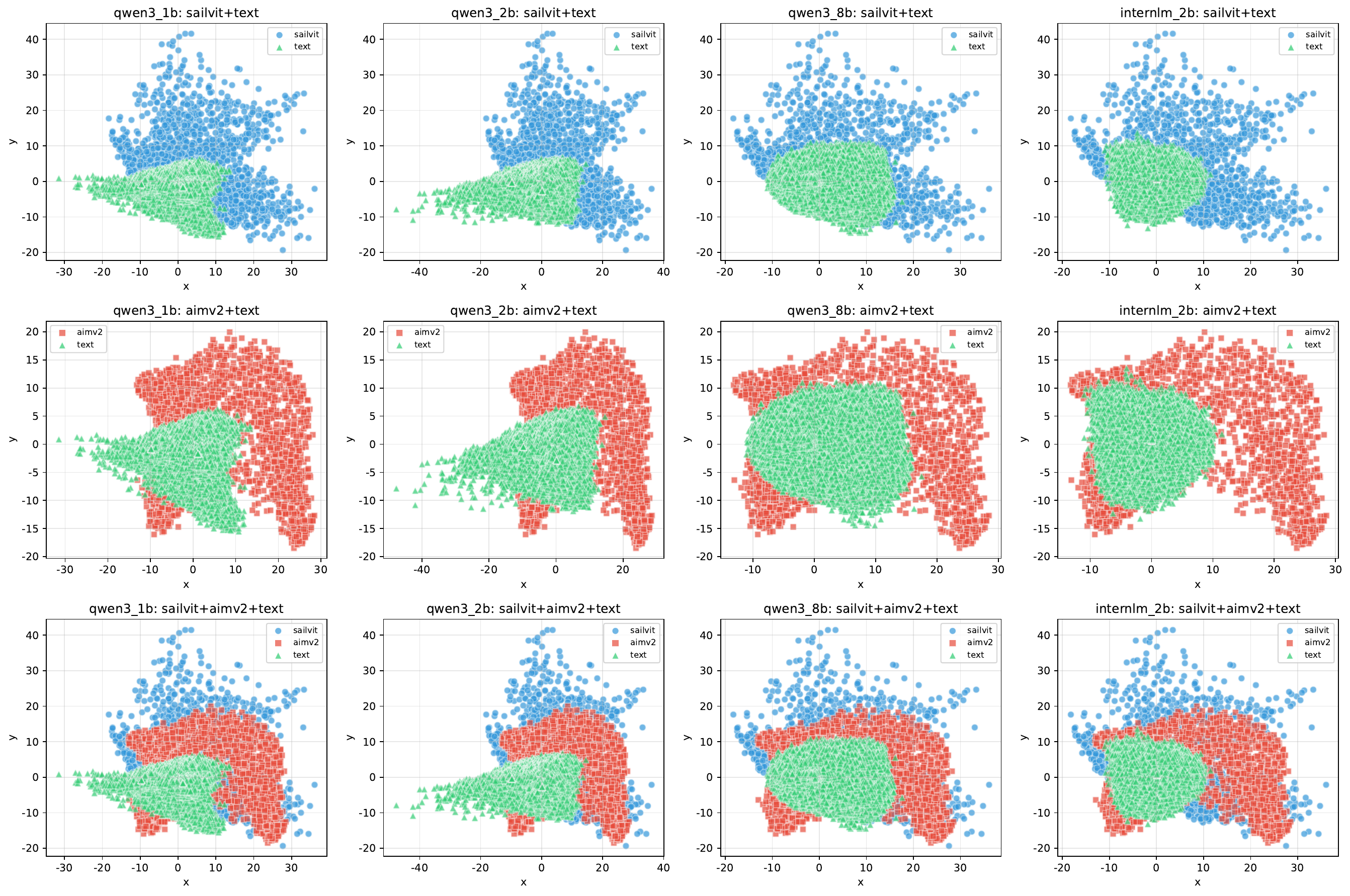}};
\end{tikzpicture}
}
\caption{
    Visualization of token embedding distributions across different models and combination strategies. Each row corresponds to a model variant (Qwen3-0.6B, Qwen3-1.7B, Qwen3-8B, InternLM-1.8B), and each column illustrates a grouping strategy: combining SAIL-ViT with text embeddings, AIMv2 with text embeddings, or all three jointly. Colors and markers indicate 'SAIL-ViT', 'AIMv2', and 'text' tokens, highlighting spatial relationships and overlaps between embedding spaces.
}
\label{fig: affinity}
\end{figure*}

\noindent\textbf{Exploring Affinity among Multimodal Features.} 
The core capability of the vision encoder is to align visual features with the LLM space. To evaluate this, we explore the affinity between multimodal features. Specifically, we calculate the distribution distance between visual features extracted by SAIL-ViT (and its baseline) and textual features from LLMs of different sizes and types. We randomly sampled five images from the internet, extracted their visual features using both SAIL-ViT and its baseline, and concatenated them to form a feature collection of size [5120, 1024]. Then, we applied Principal Component Analysis (PCA) to reduce the dimensionality to [5120, 2]. On the textual side, we extracted feature vectors from the lookup tables of different LLMs. We quantified the distribution distances by computing the average nearest neighbor distance, Wasserstein distance, and mean overall distance between these feature clusters. As shown in Figure~\ref{fig: affinity}, the visualization results show that the visual feature vectors extracted by SAIL-ViT are more compact and exhibit greater overlap with textual feature vectors, whereas the baseline model produces more dispersed visual features with less overlap with the textual space. Quantitatively, as shown in the table below, the visual feature space of SAIL-ViT is significantly closer to the textual feature space of LLMs across different sizes and architectures, as measured by multiple distance metrics. These results demonstrate that SAIL-ViT effectively reduces the gap between visual and textual feature spaces.

\begin{table}[t]
\centering
\small
\resizebox{\textwidth}{!}{%

\begin{tabular}{l cccccccc}

\toprule

\textbf{Benchmark} & \multicolumn{1}{c}{\shortstack[c]{\textbf{SAIL-VL2}\\{2B}}} & \multicolumn{1}{c}{\shortstack[c]{\textbf{SAIL-VL2}\\{Anyres-2B}}} & \multicolumn{1}{c}{\shortstack[c]{\textbf{SAIL-VL1.5}\\{2B}}} & \multicolumn{1}{c}{\shortstack[c]{\textbf{Qwen2.5-VL}\\{3B}}} & \multicolumn{1}{c}{\shortstack[c]{\textbf{Ovis-U1}\\{3B}}} & \multicolumn{1}{c}{\shortstack[c]{\textbf{InternVL3.5}\\{2B}}} & \multicolumn{1}{c}{\shortstack[c]{\textbf{InternVL3}\\{2B}}} & \multicolumn{1}{c}{\shortstack[c]{\textbf{Ovis2}\\{2B}}} \\ 

\midrule

OpenCompass$_{\text{avg}}$ & \textbf{70.31} & 68.56 & 68.08 & 65.36 & \underline{69.94} & 66.64 & 64.97 & 65.49 \\
OpenSource$_{\text{avg}}$ & \textbf{51.07} & \underline{50.15} & 47.01 & 48.40 & 42.26 & 49.49 & 46.65 & 46.53 \\ 

\midrule

\rowcolor{gray!15} \multicolumn{9}{c}{\textit{General}} \\

\midrule

MMBench$_{\text{v1.1}}$ & \textbf{86.77} & \underline{86.48} & 85.05 & 82.40 & 85.01 & 83.10 & 84.25 & 81.79 \\
MME & 76.58 & 73.36 & 73.67 & \textbf{77.46} & 76.27 & 75.99 & \underline{77.44} & 72.32 \\
MMStar & \textbf{64.07} & \underline{63.40} & 62.80 & 55.13 & 61.20 & 57.2 & 61.33 & 58.13 \\
RealWorldQA & \textbf{72.29} & 69.80 & 67.06 & 65.36 & \underline{70.07} & 60.92 & 64.58 & 66.41 \\
AI2D & 83.00 & 82.67 & \underline{83.68} & 80.73 & \textbf{85.88} & 78.63 & 78.72 & 82.80 \\
DocVQA & \underline{93.10} & 92.86 & 91.62 & 93.11 & \textbf{94.10} & 88.49 & 87.46 & 91.49 \\
OCRBench & \textbf{89.50} & 84.70 & \underline{88.50} & 83.10 & 88.10 & 83.60 & 83.40 & 86.80 \\
MMVet & 68.67 & 67.80 & 62.75 & 64.91 & \underline{69.08} & \textbf{69.45} & 66.19 & 57.34 \\
HallusionBench & \underline{51.74} & 51.60 & 49.80 & 48.29 & \textbf{55.42} & 48.16 & 41.42 & 50.05 \\
CRPE$_{\text{relation}}$ & 75.20 & \underline{75.46} & 73.92 & 72.97 & 75.15 & \textbf{75.53} & 71.58 & 73.09 \\
MMMU$_{\text{val}}$ & 47.67 & 45.00 & 44.89 & \underline{48.11} & 45.33 & \textbf{51.11} & 47.11 & 42.67 \\
MM-IFEval & \underline{52.21} & \textbf{53.26} & 34.27 & 42.68 & 46.62 & 46.04 & 33.95 & 42.09 \\ 
Refcoco$_{\text{avg}}$ & \underline{53.28} & \textbf{57.82} & 7.87 & 34.15 & 48.84 & 40.35 & 25.33 & 31.91 \\
\midrule
Overall&\textbf{70.31}&	\underline{69.55}&	63.53 &65.26 & 69.31 & 66.04 & 63.29	& 64.38 \\

\midrule

\rowcolor{gray!15} \multicolumn{9}{c}{\textit{Math \& Reasoning}} \\

\midrule

LogicVista & \underline{36.24} & 35.79 & 35.35 & 36.02 & 33.33 & \textbf{43.40} & 33.56 & 33.56 \\
MathVision & \textbf{23.36} & 19.57 & 17.60 & 17.96 & 19.24 & 19.57 & \underline{20.23} & 18.22 \\
MathVerse$_{\text{mini}}$ & 31.19 & 30.86 & 29.87 & \underline{32.97} & \textbf{37.61} & 32.87 & 32.69 & 28.27 \\
MathVista$_{\text{mini}}$ & \textbf{71.10} & 66.80 & 67.20 & 60.20 & \underline{69.50} & 61.90 & 57.30 & 64.30 \\
OlympiadBench$_{\text{mini}}$ & \underline{7.54} & 3.61 & 2.30 & 4.26 & 5.25 & \textbf{10.16} & 5.57 & 3.61 \\
WeMath & \underline{22.67} & \textbf{23.33} & 17.14 & 20.67 & 16.76 & 18.95 & 12.95 & 10.10 \\
DynaMath & 10.18 & 10.58 & 9.98 & 10.98 & \textbf{17.76} & \underline{15.17} & 13.97 & 10.78 \\
\midrule

Overall&\textbf{28.90}&	27.22&	25.63 & 26.15&	28.49& \underline{28.86}& 25.18	&24.12  \\

\midrule

\rowcolor{gray!15} \multicolumn{9}{c}{\textit{Multi-images \& Video}} \\ 

\midrule
Video-MME$_{\text{w/o\ sub}}$ & \underline{57.10} & 55.30 & 56.60 & \textbf{60.60} & 20.80 & 55.00 & 55.60 & 54.70 \\
LongVideoBench$_{\text{val}}$ & \textbf{54.45} & 53.63 & 53.18 & \underline{54.23} & 24.31 & 51.53 & 50.11 & 50.79 \\
TempCompass$_{\text{avg}}$  & 61.86 & 62.79 & 61.53 & 62.48 & 35.57 & \underline{63.20} & 62.33 & \textbf{63.47} \\
MMIU & 42.61 & \underline{43.02} & 36.18 & 37.86 & 35.19 & \textbf{44.51} & 37.88 & 35.45 \\
\midrule

Overall&\textbf{54.01}	&53.69&	51.87	&\underline{53.79}&	28.97	&53.56&	51.48	&51.10 \\

\bottomrule

\end{tabular}%
}

\caption{\textbf{Overall comparison of the SAIL-VL2 series and existing open-source MLLMs (<4B).} OpenCompass incorporates eight evaluation datasets: MMBench v1.1, MMStar, AI2D, OCRBench, MMVet, HallusionBench, MMMU${\text{val}}$, and MathVista${\text{mini}}$. The OpenSource metric is computed as the average score across three dimensions: General, Math \& Reasoning, and Multi-image \& Video. Refcoco${\text{avg}}$ is obtained by averaging five test sets: refcoco${\text{testA}}$, refcoco${\text{testB}}$, refcoco${\text{test}}$, refcocoplus${\text{testA}}$, and refcocoplus${\text{testB}}$. OlympiadBench$_{\text{mini}}$ denotes a randomly sampled subset of the original evaluation set. For the Refcoco series, prompts explicitly specified output coordinates in the 0–1000 range, with safeguards to handle outputs in the 0–1 range. Video results are reported using 16 randomly sampled frames as visual input.}
\label{tab:base-4b}
\end{table}

\begin{table}[t]
\centering

\small

\resizebox{\textwidth}{!}{%
\setlength{\tabcolsep}{3pt}

\begin{tabular}{l ccccccc cc cc}
\toprule

\textbf{Benchmark} & \multicolumn{1}{c}{\shortstack[c]{\textbf{SAIL-VL2}\\\textbf{8B}}} & \multicolumn{1}{c}{\shortstack[c]{\textbf{SAIL-VL1.6}\\\textbf{8B}}} & \multicolumn{1}{c}{\shortstack[c]{\textbf{Qwen2.5-VL}\\\textbf{7B}}} & \multicolumn{1}{c}{\shortstack[c]{\textbf{InternVL3.5}\\\textbf{8B}}} & \multicolumn{1}{c}{\shortstack[c]{\textbf{Keye-VL}\\\textbf{8B}}} & \multicolumn{1}{c}{\shortstack[c]{\textbf{InternVL3}\\\textbf{8B}}} & \multicolumn{1}{c}{\shortstack[c]{\textbf{Ovis2}\\\textbf{8B}}} & \multicolumn{1}{c}{\shortstack[c]{\textbf{SAIL-VL2}\\\textbf{A3B}}} & \multicolumn{1}{c}{\shortstack[c]{\textbf{Kimi-VL}\\\textbf{A3B}}} & \multicolumn{1}{c}{\shortstack[c]{\textbf{Seed-1.6}\\\textbf{Auto}}} & \multicolumn{1}{c}{\shortstack[c]{\textbf{GPT-4.1}}} \\

\midrule
OpenCompass$_{\text{avg}}$ & \textbf{75.07} & 74.04 & 70.62 & 73.49 & 72.77 & 73.86 & 72.75 & \underline{74.90} & {69.10} & 72.45 & 71.59 \\
OpenSource$_{\text{avg}}$ & 57.20 & 53.87 & 53.14 & 56.70 & 55.14 & 55.92 & 53.71 & 58.50 & 52.28 & \textbf{64.80} & \underline{58.72} \\ 
\midrule
\rowcolor{gray!15} \multicolumn{12}{c}{\textit{General}} \\
\midrule

MMBench$_{\text{v1.1}}$ & 90.16 & 88.54 & 86.58 & 87.59 & \textbf{93.46} & 89.68 & 88.00 & \underline{90.63} & {87.04} & 75.41 & 89.08 \\

MME & 84.54 & 81.43 & 82.21 & 85.17 & 78.48 & \underline{86.55} & 82.84 & 84.00 & {77.63} & \textbf{87.04} & 85.21 \\

MMStar & \textbf{70.73} & 69.07 & 63.93 & 68.27 & \underline{69.40} & 68.60 & 64.07 & 68.93 & {62.73} & 66.67 & 65.93 \\

RealWorldQA & \underline{76.34} & \textbf{76.60} & 67.71 & 68.10 & 65.88 & 70.59 & 72.81 & 73.59 & {67.84} & 73.07 & 71.50 \\

AI2D & \textbf{87.73} & \underline{87.47} & 84.00 & 84.03 & 85.98 & 85.17 & 86.82 & 86.63 & {84.78} & 81.83 & 83.94 \\

DocVQA & \textbf{95.28} & 94.78 & \underline{94.84} & 91.41 & 79.27 & 92.03 & 94.21 & 93.84 & {98.70} & 91.51 & 63.24 \\

OCRBench & \textbf{91.30} & 90.40 & 87.80 & 83.80 & 81.10 & 88.30 & 88.90 & \underline{90.60} & {86.00} & 80.30 & 74.20 \\ 

MMVet & \underline{73.72} & 74.36 & 70.05 & 80.50 & 67.20 & 82.71 & 70.96 & \textbf{74.36} & {65.83} & 72.57 & 73.30 \\

HallusionBench & 55.10 & 54.53 & 55.97 & 53.59 & \textbf{57.78} & 49.00 & 55.92 & 52.81 & {46.54} & 50.05 & \underline{56.12} \\

CRPE$_{\text{relation}}$ & \underline{77.61} & 77.24 & 76.06 & 75.01 & 76.74 & 76.58 & 77.02 & 76.72 & {75.00} & \textbf{80.37} & 77.52 \\

MMMU$_{\text{val}}$ & 55.44 & 52.33 & 50.33 & 57.22 & 59.11 & 56.78 & 56.44 & 60.56 & {53.67} & \textbf{76.44} & \underline{63.56} \\

MM-IFEval & 60.81 & 52.25 & 50.78 & 55.86 & 54.67 & 53.93 & 53.96 & 61.07 & {43.33} & \underline{72.10} & \textbf{72.23} \\ 

Refcoco$_{\text{avg}}$ & 74.02 & 37.25 & 34.19 & \textbf{88.89} & 48.08 & \underline{80.93} & 31.91 & 79.28 & - & 79.67 & 11.13\\

\midrule

Overall& \underline{76.37} & 72.02 & 69.57 & 75.34 & 70.55 & 75.45 & 71.07 & \textbf{76.39}  & 70.76 & 75.92 & 68.23 \\

\midrule

\rowcolor{gray!15} \multicolumn{12}{c}{\textit{Math \& Reasoning}} \\ 
\midrule

LogicVista & 44.97 & 42.51 & 40.72 & 52.57 & 43.18 & 45.41 & 41.16 & 47.20 & {42.70} & \textbf{63.76} & \underline{54.36} \\

MathVision & 27.63 & 22.11 & 23.59 & 21.97 & 17.11 & 28.62 & 25.92 & 28.55 & {17.04} & \textbf{67.24} & \underline{39.14} \\

MathVerse$_{\text{mini}}$ & 43.17 & 40.76 & 43.15 & 35.86 & 47.34 & 40.84 & 36.50 & 44.53 & {40.61} & \textbf{70.94} & \underline{51.14} \\

MathVista$_{\text{mini}}$ & \textbf{76.40} & 75.60 & 66.30 & 72.90 & 68.10 & 70.60 & 70.90 & 74.70 & {64.50} & \underline{76.30} & 66.60 \\

OlympiadBench$_{\text{mini}}$ & 14.10 & 6.23 & 7.21 & 16.07 & 15.41 & 12.79 & 7.87 & \underline{16.39} & {8.85} & \textbf{26.56} & 14.43 \\

WeMath & 35.81 & 27.62 & 32.19 & 31.81 & 48.57 & 31.62 & 26.57 & 31.62 & {34.00} & \textbf{52.76} & \underline{49.52} \\

DynaMath & 17.76 & 14.77 & 14.37 & 20.76 & 23.15 & 16.37 & 20.16 & 19.76 & {18.36} & \textbf{39.92} & \underline{37.13} \\ 

\midrule

Overall& 37.12 & 32.80 & 32.50 & 35.99 & 37.55 & 35.18 & 32.73 & 37.54 & {32.29} & \textbf{56.78} & \underline{44.62} \\

\midrule
\rowcolor{gray!15} \multicolumn{12}{c}{\textit{Multi-images \& Video}} \\ 
\midrule

Video-MME$_{\text{w/o\ sub}}$ & 62.70 & 61.80 & 64.80 & 62.00 & 59.10 & 62.10 & 62.20 & 65.10 & {66.00} & \underline{66.00} & \textbf{69.50} \\

LongVideoBench$_{\text{val}}$ & 58.34 & 57.07 & \textbf{59.69} & 56.84 & 57.20 & 55.20 & 56.39 & \underline{59.54} & {59.40} & 49.74 & 57.44 \\
 
TempCompass$_{\text{avg}}$  & 65.66 & 67.27 & 63.74 & 69.42 & 67.02 & 69.97 & 71.01 & 67.72 & {48.28} & \underline{73.59} & \textbf{76.67} \\

MMIU & 45.72 & 40.97 & 41.15 & 46.85 & 45.97 & 41.25 & 39.76 & \underline{53.89} & {41.50} & \textbf{57.50} & 49.60 \\
\midrule

Overall&58.11&	56.78	&57.35	& 58.78 & 57.32& 57.13 & 57.34 & 61.56 & {53.79} & \underline{61.71} & \textbf{63.30}\\

\bottomrule
\end{tabular}%
}

\caption{\textbf{Overall comparison of the SAIL-VL2 series with existing open-source 8B MLLMs and closed-source models.} OpenCompass includes eight evaluation datasets: MMBench v1.1, MMStar, AI2D, OCRBench, MMVet, HallusionBench, MMMU${\text{val}}$, and MathVista${\text{mini}}$. The OpenSource metric is the average score across three dimensions: General, Math \& Reasoning, and Multi-image \& Video. Refcoco${\text{avg}}$ is computed over five test sets: refcoco${\text{testA}}$, refcoco${\text{testB}}$, refcoco${\text{test}}$, refcocoplus${\text{testA}}$, and refcocoplus${\text{testB}}$. OlympiadBench$_{\text{mini}}$ denotes a randomly sampled subset of the original benchmark. For Refcoco, prompts specify output coordinates in the 0–1000 range, with safeguards for outputs in 0–1. Video results are obtained by sampling 16 frames. Results for Kimi-VL-A3B are reported in no-thinking mode; Seed-1.6 Auto denotes auto-thinking mode. Kimi-VL-A3B exhibits poor instruction-following on the customized Refcoco set, yielding unusable outputs, which are marked as “–” and excluded from metric calculations.}
\label{tab:base-8b}
\end{table}

\subsubsection{Multimodal Understanding Tasks}


\noindent\textbf{General Multimodal Understanding.}
To comprehensively evaluate the general multimodal understanding capabilities of SAIL-VL2 across multiple dimensions, including visual question answering, document understanding, and OCR, we conduct extensive experiments on a diverse range of benchmarks. 

As reported in Table~\ref{tab:base-4b} and ~\ref{tab:base-8b}, SAIL-VL2-2B and SAIL-VL2-8B achieve state-of-the-art performance across diverse benchmarks covering multiple evaluation dimensions. At comparable parameter scales, both models consistently outperform prior open-source counterparts of the same size (2B/8B) on representative datasets, including the MMBench series~\citep{liu2024mmbench}, MMStar~\citep{liu2024mmbench}, RealWorldQA~\citep{real_world_qa}, DocVQA~\citep{mathew2021docvqadatasetvqadocument}, and OCRBench~\citep{ocrbench}. These results highlight the superior general multimodal understanding capabilities of SAIL-VL2.

In the domain of general visual question answering, SAIL-VL2-2B achieves leading performance, obtaining 86.77 on MMBench-v1.1, 72.29 on RealWorldQA, and 64.07 on MMStar. These results surpass previous sub-4B state-of-the-art LVMs, including Ovis-U1, Qwen2.5-VL-3B, and InternVL3.5-2B, establishing SAIL-VL2-2B as a benchmark efficient model for visual detail understanding and demonstrating the feasibility of achieving high performance with compact architectures.

In the domain of document image understanding, SAIL-VL2 achieves consistently strong results across OCR and document comprehension tasks. On OCRBench, SAIL-VL2-2B attains a score of 895, setting the state of the art among models below 4B parameters, while SAIL-VL2-8B further improves to 913, ranking first within the 8B scale. On DocVQA, SAIL-VL2-2B reaches 93.10, outperforming all models of the same size and remaining competitive with larger 3B models, whereas SAIL-VL2-8B achieves 95.28, establishing clear superiority among 8B-scale models. These results highlight SAIL-VL2’s leading capability in text–image understanding, demonstrating both efficiency and scalability across model sizes.

\begin{table}[t] 
\centering 
\small 
\resizebox{\textwidth}{!}{%
\setlength{\tabcolsep}{4pt} 
\begin{tabular}{lcccccc|c} 
\toprule 
\textbf{Model} & \textbf{MathVista} & \textbf{MathVision} & \textbf{MathVerse} & \textbf{DynaMath} & \textbf{WeMath} & \textbf{LogicVista} & \textbf{Average} \\ 

\midrule 
\rowcolor{gray!15} \multicolumn{8}{c}{\textbf{\textit{Close-source Models}}} \\ 
\midrule 

Gemini-2.0-Flash & 70.4 & 47.8 & 43.6 & 42.1 & 47.4 & 52.3 & 50.6 \\ 
Gemini-2.0-Pro & 71.3 & 48.1 & 67.3 & 43.3 & 56.5 & 53.2 & 56.6 \\ 
GPT-4.1-20250414 & 70.4 & 45.1 & 48.9 & 43.3 & 55.5 & 61.1 & 54.0 \\ 
GPT-4o-latest & 71.6 & 43.8 & 49.9 & 48.5 & 50.6 & 64.4 & 54.8 \\ 
Claude-3.7-Sonnet & 66.8 & 41.3 & 52.0 & 39.7 & 58.2 & 49.3 & 49.6 \\ 

\midrule 
\rowcolor{gray!15} \multicolumn{8}{c}{\textbf{\textit{Open-source Models}}} \\ 
\midrule 

OVR-7B & 72.1 & \underline{51.8} & 54.6 & 33.5 & 44.6 & 54.8 & {51.9} \\ 
WeThink-7B & 70.9 & 27.2 & 44.7 & 24.4 & 48.0 & 53.0 & 44.7 \\ 
Qwen2.5-VL-7B & 68.1 & 25.4 & 41.1 & 21.8 & 36.2 & 47.9 & 40.1 \\ 
Qwen2.5-VL-72B & 74.2 & 39.3 & 47.3 & \underline{35.9} & {49.1} & {55.7} & 50.2 \\ 
InternVL3-8B & 70.5 & 30.0 & 38.5 & 25.7 & 39.5 & 44.5 & 41.4 \\ 
InternVL3-78B & \underline{79.0} & 38.8 & 51.0 & {35.1} & 46.1 & {55.9} & 51.0 \\ 
VL-Rethinker-7B & 73.7 & 28.4 & 46.4 & 17.8 & 36.3 & 42.7 & 40.9 \\ 
VLAA-Thinker-7B & 68.0 & 26.4 & 48.2 & 22.4 & 41.5 & 48.5 & 42.5 \\ 
OpenVLThinker-7B & 65.3 & 23.0 & 38.1 & 16.8 & 35.2 & 44.5 & 37.2 \\ 
Keye-VL-8B-Thinking & {77.2} & 43.7 & 53.4 & \textbf{37.1}  & \textbf{60.2} & 49.2 & 53.5 \\ 
Kimi-VL-A3B-Thinking-2506 & \textbf{79.5} & \textbf{53.6} & {55.2} & 29.1 & 45.4 & 47.2 & 51.7 \\ 
\midrule 
SAIL-VL2-2B-Thinking & 68.5 & 27.5 & 43.4 & 20.2 & 38.8 & 47.0 & 40.9 \\ 
SAIL-VL2-8B-Thinking & 75.8 & 46.7 & \textbf{58.9} & 33.5 & \underline{54.9} & \underline{56.4} & \textbf{54.4} \\ 
SAIL-VL2-A3B-Thinking & 73.0 & 44.9 & \underline{55.7} & 34.1 & {54.2} & \textbf{59.7} & \underline{53.6} \\ 
\bottomrule 
\end{tabular}%
} 
\caption{Evaluation results on OpenCompass multimodal reasoning benchmarks. All the results are reported from OpenCompass with the GPT-4o-Mini as the judge model. The best results among open-source models are \textbf{bolded} and the second-best results are \underline{underlined}.} 
\label{reasoning_vl} 
\end{table}

\noindent\textbf{Visual Grounding.} 
We further assess SAIL-VL2 on fine-grained visual understanding, with a focus on visual grounding. On RefCOCO~\citep{refcoco}, SAIL-VL2-AnyRes-2B, equipped with an arbitrary-resolution vision encoder, achieves 57.82 on RefCOCO$_{avg}$. This represents a clear improvement over the fixed-resolution SAIL-VL2-2B and establishes state-of-the-art performance among sub-4B LVMs. These results demonstrate the effectiveness of incorporating resolution-adaptive encoders and highlight the flexibility of the SAIL-ViT family in adapting to fine-grained visual understanding tasks.

\noindent\textbf{Multi-images and Video Understanding.} We further evaluate the multi-image and video understanding capabilities of SAIL-VL2 across diverse benchmarks. In the domain of video understanding, SAIL-VL2 achieves leading results on the general-purpose Video-MME benchmark~\citep{video_llava}, where SAIL-VL2-2B attains 57.10. On the long-video benchmark LongVideoBench~\citep{longvideobench}, SAIL-VL2-2B establishes state-of-the-art performance among models below 4B parameters, demonstrating strong ability to handle extended temporal contexts. For multi-image reasoning, SAIL-VL2-2B reaches 42.61 on the MMIU benchmark~\citep{hendryckstest2021}, again achieving leading performance. These results collectively verify SAIL-VL2’s comprehensive strengths in both multi-image and video understanding, spanning short to long temporal horizons.

\subsubsection{Multimodal Reasoning Tasks}

We evaluate our SAIL-VL2-thinking models on the OpenCompass multimodal reasoning benchmarks to assess their advanced cognitive abilities. This benchmark was chosen because it comprises a diverse set of challenging tasks designed to measure deep reasoning rather than simple object recognition. The included tasks are MathVista~\citep{lu2023mathvista}, MathVerse~\citep{mishra2024mathverse}, MathVision, LogicVista, WeMath~\citep{fang2024wemath}, and DynaMath. Together, these tests cover a wide range of skills, including visual mathematics, which requires solving math problems presented in images, complex logical reasoning, and dynamic problem-solving.

As shown in Table~\ref{reasoning_vl}, our models achieve leading results on the OpenCompass benchmark. Our premier dense model, SAIL-VL2-8B-Thinking , establishes a new state-of-the-art for open-source models by securing the top position on the leaderboard with a score of 54.4. Our Mixture-of-Experts (MoE) and smaller dense models also demonstrate exceptional efficiency and performance. Notably, our SAIL-VL2-MoE-Thinking model achieves a high score of 53.6 while using only 3B activated parameters. This result is significant as it shows a strong balance between performance and computational cost. In terms of performance, this score not only surpasses strong closed-source models like Gemini-2.0-Flash (50.6) but is also highly competitive with the top-tier GPT-4o-latest (54.8). It also holds a clear advantage over other open-source MoE models, such as kimi-vl-A3B-thinking-2506 (51.7). Taken together, these results, combined with the strong performance of our dense SAIL-VL2-A3B-Thinking model, validate the overall effectiveness and scalability of our SAIL-VL2-Thinking architecture, confirming its successful implementation in both MoE and dense model configurations.

\section{Conclusion}
In this report, we present SAIL-VL2, an open-suite vision–language foundation model for comprehensive multimodal understanding and reasoning. Through innovations in data curation, progressive training, and architecture, SAIL-VL2 achieves consistent gains in both efficiency and performance. Across 106 benchmarks, it sets state-of-the-art results at the 2B and 8B scales. Notably, SAIL-VL2-2B delivers leading results on MMMU and MathVista, and ranks first on the OpenCompass leaderboard among officially released models under 3B, underscoring its competitiveness as an efficient yet powerful LVM. These findings establish SAIL-VL2 as both a high-performance milestone and a scalable foundation for the open-source multimodal community. Looking forward, we will further enhance the SAIL-VL series through more efficient architectures, comprehensive pre-training strategies, and improved reinforcement learning paradigms, enabling its continuous evolution toward stronger multimodal intelligence.

\section{Contributors and Acknowledgment}
All contributors are listed in reverse alphabetical order by last name initial, with equal contributions within each group; unless otherwise noted, all are members of the Douyin SAIL Team.

\textbf{Core Contributors:} \\
Weijie Yin \quad Yongjie Ye \quad Fangxun Shu \quad Yue Liao~(LV-NUS) \quad Zijian Kang \quad Hongyuan Dong

\textbf{Contributors:} \\
Haiyang Yu \quad Dingkang Yang \quad Jiacong Wang \quad Han Wang \quad Wenzhuo Liu

\textbf{Project Leader:} \\
Xiao Liang

\textbf{Supervisors:} \\
Xiao Liang \quad Shuicheng Yan~(LV-NUS) \quad Chao Feng

\textbf{Acknowledgement:} \\
We gratefully acknowledge the foundational contributions of prior works, including SAIL-ViT~\citep{yin2025sailvit} and AdaLRS~\citep{dong2025adalrs}. We also sincerely thank Yuqing Zhou, Sheng Zheng, Xiao Yu, Yuzhong Wang, Yan Qiu, Yifan Pi, Yaling Mou, Changyue Liao, Yuzhuo Li, Yifan Jiang, Xinjie Huang, and Zirui Guo for their valuable support and contributions.

\bibliography{colm2024_conference}
\bibliographystyle{colm2024_conference}

\newpage
\appendix

\end{document}